\documentclass[sigconf, nonacm]{acmart}

\AtBeginDocument{}
\acmYear{2026}
\acmBooktitle{Manuscript draft}
\acmDOI{}
\makeatletter
\let\recog@mkbibcitation\@mkbibcitation
\def\@mkbibcitation{\@ACM@nonacmfalse\recog@mkbibcitation\@ACM@nonacmtrue}
\makeatother

\usepackage{booktabs}
\usepackage{float}
\floatstyle{ruled}
\newfloat{algorithm}{t}{loa}
\floatname{algorithm}{Algorithm}
\usepackage{flushend}
\usepackage{multirow}
\usepackage{graphicx}
\usepackage{xcolor}
\ifdefined\symbf
  \let\bm\symbf
\else
  \usepackage{bm}
\fi
\usepackage{enumitem}
\usepackage{tikz}
\usetikzlibrary{arrows.meta,positioning,calc,fit,backgrounds,decorations.pathreplacing}

\newcommand{\recog}{ReCoG}
\newcommand{\Att}{\mathrm{att}}
\newcommand{\Str}{\mathrm{str}}
\newcommand{\Sem}{\mathrm{sem}}

\begin{document}
\settopmatter{printacmref=true, authorsperrow=4}

\title{ReCoG: Relative Node Contexts for Multi-Domain Graph Pre-training}

\author{Yupeng Tong}
\authornotemark[2]
\email{cualb@tju.edu.cn}
\affiliation{%
  \department{School of Computer Science and Technology}
  \institution{Tianjin University}
  \city{Tianjin}
  \country{China}}
\author{Lianze Shan}
\authornotemark[2]
\email{shanlz2119@tju.edu.cn}
\affiliation{%
  \department{School of Computer Science and Technology}
  \institution{Tianjin University}
  \city{Tianjin}
  \country{China}}
\author{Jitao Zhao}
\email{zjtao@tju.edu.cn}
\affiliation{%
  \department{School of Computer Science and Technology}
  \institution{Tianjin University}
  \city{Tianjin}
  \country{China}}
\author{Dongxiao He}
\authornotemark[1]
\email{hedongxiao@tju.edu.cn}
\affiliation{%
  \department{School of Computer Science and Technology}
  \institution{Tianjin University}
  \city{Tianjin}
  \country{China}}

\begin{abstract}
The success of foundation models in language and vision has motivated Graph Foundation Models (GFMs), which learn transferable knowledge through multi-domain graph pre-training. Existing methods advance GFMs through feature unification and structural learning, and in doing so they rely on the raw graph structure to organize node contexts. However, we observe that removing a large proportion of edges from source graphs during pre-training improves cross-domain transfer accuracy across four representative methods. Our theoretical analysis identifies the raw graph structure as a source of conflict in cross-graph model sharing. We therefore advocate defining node contexts using criteria consistent across graphs. Achieving this requires (1) constructing node contexts consistently across graphs without a shared positional structure, such as the token order of text or the pixel grid of images, and (2) retaining the structural and neighborhood information of the original graphs. To this end, we introduce \textbf{Re}lative \textbf{Co}ntexts for \textbf{G}raphs (\textbf{ReCoG}) for multi-domain graph pre-training. Specifically, we first construct attribute contexts to provide a comparison rule independent of the raw graph structure. Furthermore, we construct structural contexts based on common neighbors and semantic contexts based on neighborhood attribute summaries to retain the topological and neighborhood information of the original graphs. Since all graphs share these construction rules, we pre-train a single graph encoder by reconstructing node attributes over these contexts, without domain-specific modules. We conduct extensive experiments on 18 target datasets, which demonstrate the effectiveness of ReCoG. In particular, ReCoG exceeds the strongest baseline by 7.61 percentage points in mean 1-shot cross-domain node classification accuracy and by 4.47 points in cross-type transfer to heterophilous graphs.
\end{abstract}

\ccsdesc[500]{Computing methodologies~Learning latent representations}
\ccsdesc[500]{Information systems~Web mining}
\ccsdesc[300]{Information systems~Data mining}

\keywords{Graph Neural Networks, Graph Foundation Models, Graph Pre-training}

\maketitle
\begingroup
\renewcommand{\thefootnote}{\fnsymbol{footnote}}
\footnotetext[2]{These authors contributed equally to this work.}
\footnotetext[1]{Corresponding author.}
\endgroup

\section{Introduction}

\label{sec:intro}

\begin{figure}[t]

\centering

\includegraphics[width=1.0\linewidth,trim=0 4bp 0 4bp,clip]{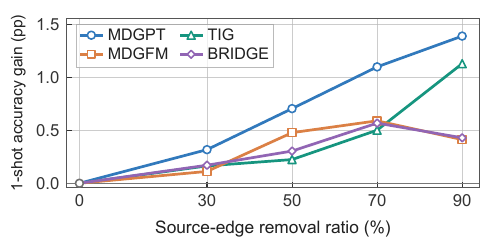}

\caption{Mean 1-shot accuracy gains (percentage points) across six unseen target graphs under source-edge removal, relative to the same model pre-trained on full source graphs.}

\Description{Line chart of mean one-shot accuracy gains for MDGPT, TIG, MDGFM, and BRIDGE at source-edge removal ratios of zero, thirty, fifty, seventy, and ninety percent. MDGPT and TIG rise across the tested ratios, while MDGFM and BRIDGE peak at seventy percent. All four have positive gains at each nonzero removal ratio.}

\label{fig:source_edge_deletion}

\end{figure}

Graphs represent entities and their relationships across the Web, including citation networks~\cite{sen2008collective}, recommender systems~\cite{ying2018graph}, and knowledge graphs, while also organizing retrieval in retrieval-augmented generation~\cite{edge2024local,he2024gretriever} and modeling communication and task dependencies among large language model (LLM) agents~\cite{zhuge2024gptswarm,zhang2025gdesigner}. In language and vision, foundation models pre-trained on large-scale corpora generalize broadly to new data and downstream tasks~\cite{bommasani2021opportunities,brown2020language,dosovitskiy2021image}. Extending this paradigm to graphs is challenging because graphs from different applications vary widely in attribute dimensions, connectivity patterns, and task objectives. Graph Foundation Models (GFMs) address this challenge through multi-domain graph pre-training, learning transferable representations from unlabeled source graphs before adapting to unseen target graphs and tasks~\cite{mao2024position,liu2025graph}.

Existing multi-domain graph pre-training methods reconcile cross-graph heterogeneity along two directions: feature alignment and structural transfer. For feature alignment, FUG~\cite{zhao2024fug} learns transformations across varying input dimensions, TIG~\cite{zhao2026towards} converts feature dimensions and their interactions into structural representations, MDGPT~\cite{yu2024text} aligns source features via domain tokens, and BRIDGE~\cite{yuan2025much} pairs domain-invariant alignment with target adaptation. For structural transfer, SAMGPT~\cite{yu2025samgpt} introduces structure tokens to modulate cross-domain aggregation, GFT~\cite{wang2024gft} constructs a vocabulary of computation trees, and MDGFM~\cite{wang2025multidomain} refines topologies using joint feature and structural signals. These methods address cross-graph heterogeneity through diverse strategies, and all of them organize node contexts by the raw graph structure. However, we observe a counterintuitive phenomenon. We randomly remove varying proportions of edges from source graphs during pre-training while keeping all other settings unchanged. As shown in Figure~\ref{fig:source_edge_deletion}, removing 90\% of these edges improves each of the four representative methods' mean 1-shot accuracy over six unseen target graphs. Our theoretical analysis identifies the raw graph structure as a source of conflict in cross-graph model sharing (Section~\ref{sec:theory}).

\begin{figure}[t]

\centering

\includegraphics[width=0.9\linewidth]{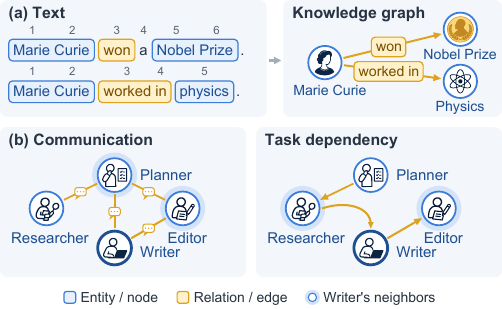}

\caption{Recorded relations determine node contexts. (a) The same facts expressed as text and labeled graph edges. (b) Communication and task dependencies give the same agents different neighbors.}

\Description{Comparison of context relations in text, a knowledge graph, and two graphs of language-model agents with different recorded relations.}

\label{fig:motivation}

\end{figure}

The raw graph structure defines a node's context by specifying its neighbors and the relations through which it aggregates their information. In language and vision, sequential token order and regular image grids provide consistent positional rules for organizing local contexts across samples. General graphs lack any shared node ordering or coordinate system; instead, the raw graph structure defines neighborhoods according to the domain-specific relations recorded during graph construction. For instance, an edge in a knowledge graph encodes a specific semantic predicate (Figure~\ref{fig:motivation}(a)), whereas the same group of LLM agents forms distinct neighborhood topologies depending on whether edges represent communication relations or task dependencies (Figure~\ref{fig:motivation}(b)). Propagating along the raw graph structure therefore ties message passing to the specific relation types recorded in each source domain. We therefore propose constructing node contexts from relative node associations defined by criteria consistent across graphs.

Realizing this approach presents two challenges. First, graphs lack a shared positional structure such as the token order of text or the pixel grid of images, so node contexts must be constructed consistently across graphs through a relational criterion that compares nodes within each graph despite disparate node orderings and recorded edge types. Second, contexts constructed by such criteria depart from the raw graph structure, yet they must retain its structural and neighborhood information, because the raw graph structure both encodes local connection patterns and determines which neighbors contribute attributes to each node.

Based on this perspective, we propose \textbf{Re}lative \textbf{Co}ntexts for \textbf{G}raphs (\textbf{ReCoG}) for multi-domain graph pre-training. Specifically, ReCoG first constructs attribute contexts that link related nodes independently of the raw graph structure, addressing the first challenge. Furthermore, to address the second challenge, it constructs structural contexts from common neighbors and semantic contexts from similarities between neighborhood attribute summaries. Message passing thus operates over these contexts, and the raw graph structure is used only to construct them. During pre-training, ReCoG trains a shared attribute encoder and GNN over the union of the three contexts via node attribute reconstruction. For downstream tasks, we freeze the pre-trained encoder, adjust the propagation weights of the three context types, and train task-specific predictors for node-, edge-, and graph-level tasks.

Our main contributions are as follows:

\begin{itemize}[leftmargin=1.2em,itemsep=1pt,topsep=2pt]

\item We identify the raw graph structure as a source of conflict in multi-domain graph pre-training. Removing a large proportion of edges from source graphs improves cross-domain transfer for four representative methods, and our theoretical analysis quantifies the conflicting reconstruction demands that graph-dependent attribute mixing imposes on a shared model.

\item We introduce ReCoG, which constructs attribute, structural, and semantic contexts under criteria consistent across graphs and adapts their propagation weights with a frozen encoder. Our analysis bounds how weighted-context perturbations affect propagation and reconstruction loss.

\item We evaluate ReCoG across 18 target datasets. It achieves leading average accuracy in cross-domain and cross-type transfer, and generalizes to node-, edge-, and graph-level tasks with a frozen encoder.

\end{itemize}

\begin{figure*}[t]
\centering
\includegraphics[width=0.92\textwidth]{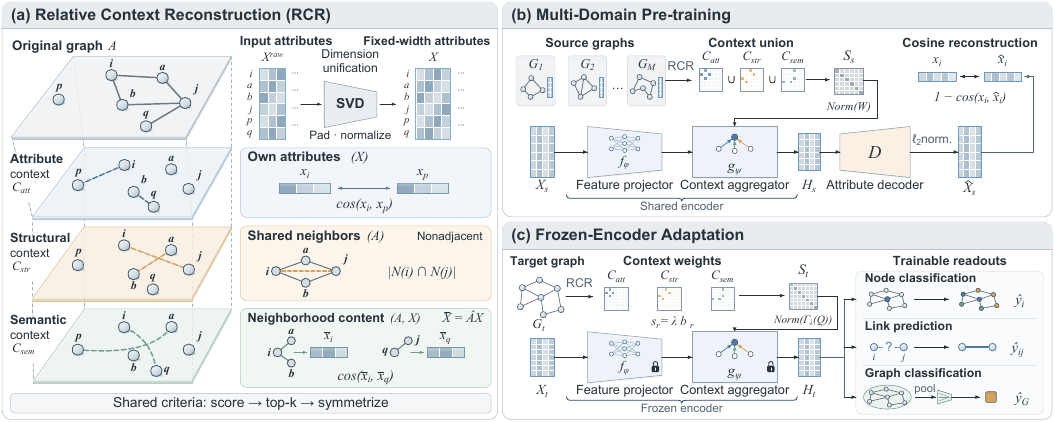}
\caption{Overview of \recog{}. (a) Shared criteria construct and symmetrize attribute, structural, and semantic contexts. (b) A shared feature projector (MLP) and context aggregator (GNN) reconstruct attributes using a deduplicated context union. (c) Context weights $s_r=\lambda b_r$ configure target propagation; the frozen encoder supports node classification, link prediction, and graph classification through task-specific predictors.}
\Description{The layered diagram uses blue, ochre and green for attribute, structural and semantic contexts. Projection guides join corresponding plane corners and fade beneath translucent surfaces; thin side faces and contact shadows indicate depth. Source graphs are visited in turn and share feature projection, context aggregation and attribute decoding. A distinct seven-node target graph uses configurable context weights and the frozen encoder. Three readout rows show classification of a selected node, prediction of a candidate link, and classification of a whole graph after pooling.}
\label{fig:framework}
\end{figure*}

\section{Related Work}
\label{sec:related}
\textbf{Graph self-supervised learning.} Contrastive and generative objectives learn graph representations without manual labels: DGI~\cite{velickovic2019dgi} contrasts node representations with graph summaries, GRACE~\cite{zhu2020deep} aligns augmented views, and BGRL~\cite{thakoor2022largescale} bootstraps representations across views. GraphMAE~\cite{hou2022graphmae} reconstructs masked node attributes, GCC~\cite{qiu2020gcc} uses subgraph discrimination to transfer structural patterns across networks, and Hu et al.~\cite{hu2020strategies} combine node- and graph-level pre-training strategies. These objectives use graph neighborhoods and their augmentations; \recog{} performs multi-domain attribute reconstruction over relative contexts defined by comparison rules shared across graphs.

\textbf{Graph foundation models.} Source graphs differ in attributes and topology. OFA~\cite{liu2024one}, ZeroG~\cite{li2024zerog}, and UniGraph~\cite{he2025unigraph} map text-attributed graphs into a shared language-model space. For general attributes, FUG~\cite{zhao2024fug} learns dimension-adaptive basis transformations, TIG~\cite{zhao2026towards} models interactions among feature dimensions, and STAGE~\cite{pmlr-v267-shen25p} encodes statistical dependencies between attributes. GCOPE~\cite{zhao2024all} connects source graphs through coordinator nodes, MDGPT~\cite{yu2024text} introduces domain tokens, LEDA~\cite{shan2026leda} aligns latent semantic distributions, and BRIDGE~\cite{yuan2025much} combines feature aligners with target-specific routing. For structural transfer, SAMGPT~\cite{yu2025samgpt} modulates aggregation via structure tokens, GFT~\cite{wang2024gft} learns a computation-tree vocabulary, GIT~\cite{pmlr-v267-wang25eq} pre-trains on task-trees, RiemannGFM~\cite{sun2025riemanngfm} learns a structural vocabulary in Riemannian geometry, MDGFM~\cite{wang2025multidomain} aligns features with refined topology, and MDGMIX~\cite{mdgmix2026} mixes source subgraphs across domain boundaries. ProGFM~\cite{wang2026beyond} learns propagation patterns between edges and feature dimensions. To adapt to target graphs, MDGPT~\cite{yu2024text} and SAMGPT~\cite{yu2025samgpt} tune target prompts, GraphControl~\cite{zhu2024graphcontrol} adds target-specific conditional control to a pre-trained model, BRIDGE~\cite{yuan2025much} routes target features through a mixture of experts, GraphLoRA~\cite{yang2025graphlora} applies structure-aware low-rank updates, and GraphAny~\cite{zhao2025fully} combines analytical linear graph predictors through learned attention for training-free target inference. \recog{} instead defines propagation through relative contexts using the same comparison rules across graphs, and adapts context weights and fits task predictors with the encoder frozen.

\textbf{Relational construction for graph learning.} IDGL~\cite{chen2020iterative} jointly refines graph connectivity and node embeddings, while SLAPS~\cite{fatemi2021slaps} uses self-supervision to learn graph connections. To handle non-local or heterophilous relations, Geom-GCN~\cite{pei2020geom} aggregates over geometric neighborhoods and H2GCN~\cite{zhu2020beyond} separates ego representations from higher-order neighborhoods. These methods learn or select neighborhoods within a single graph, whereas \recog{} applies fixed ranking rules to every source and target graph, so one frozen encoder receives contexts built under the same criteria.
\section{Preliminaries}
\label{sec:prelim}
We use calligraphic letters for sets, bold uppercase letters for matrices, and bold lowercase letters for vectors.

\textbf{Graphs and propagation.} Let $G=(\mathcal V,\mathcal E)$ be an undirected graph with $n$ nodes, raw edge set $\mathcal E$, adjacency matrix $\mathbf A$, and raw neighborhood $\mathcal N(i)$ for node $i$. We call $\mathcal E$ and $\mathbf A$ the raw graph structure. For a nonnegative symmetric weight matrix $\mathbf W$, define $\operatorname{Norm}(\mathbf W)=\bm\Delta^{-1/2}(\mathbf W+\mathbf I)\bm\Delta^{-1/2}$ with $\bm\Delta=\operatorname{diag}\bigl((\mathbf W+\mathbf I)\mathbf 1\bigr)$, and write $\widehat{\mathbf A}=\operatorname{Norm}(\mathbf A)$. Given a propagation operator $\mathbf S\in\mathbb R^{n\times n}$, a Graph Convolutional Network (GCN) layer~\cite{kipf2017semisupervised} computes
\begin{equation}
\mathbf H'=\sigma\bigl(\mathbf S\mathbf H\bm\Theta+\mathbf1\mathbf b^\top\bigr),
\label{eq:gcn}
\end{equation}
where $\bm\Theta$ and $\mathbf b$ are learnable parameters and $\sigma$ is a nonlinear activation. A standard GCN sets $\mathbf S=\widehat{\mathbf A}$, aggregating each node's representation with those of its raw neighbors.

\textbf{Multi-domain pre-training.} Given $M$ unlabeled source graphs $\{G_s\}_{s=1}^{M}$ from distinct domains, each graph $G_s$ has $n_s$ nodes and raw attributes $\mathbf X_s^{\mathrm{raw}}\in\mathbb R^{n_s\times d_s}$ whose dimension $d_s$ and semantics vary across domains. Following text-free multi-domain pre-training~\cite{yu2025samgpt,yu2024text}, we project $\mathbf X_s^{\mathrm{raw}}$ via singular value decomposition (SVD) onto up to $d$ leading right singular vectors, zero-pad the result to $d$ columns, and normalize each nonzero row to unit length. This yields $\mathbf X_s\in\mathbb R^{n_s\times d}$ with row norms at most one; we omit the subscript $s$ when clear from context and write $\mathbf x_i$ for the $i$-th row of $\mathbf X$. The encoder is pre-trained on $\{G_s\}_{s=1}^{M}$ and transferred to unseen targets with the same label-free preprocessing.

\section{Method}
\label{sec:method}
As illustrated in Figure~\ref{fig:framework}, \recog{} replaces raw-edge propagation with relative node contexts across pre-training and adaptation. Relative Context Reconstruction (RCR) first extracts attribute, structural, and semantic contexts under criteria shared across domains (Section~\ref{sec:contexts}). We then pre-train a shared encoder over the union of these contexts via node attribute reconstruction (Section~\ref{sec:pretrain}), and adapt to target graphs by reweighting the three context types while keeping the encoder frozen (Section~\ref{sec:deploy}).

\subsection{Relative Context Reconstruction}
\label{sec:contexts}
RCR defines propagation pathways by ranking node pairs under three criteria: attribute similarity provides a comparison rule independent of raw adjacency, while common neighbors and neighborhood attribute summaries preserve topological and local semantic evidence from the raw graph. Using the preprocessed attributes $\mathbf X$ and raw adjacency $\mathbf A$ from Section~\ref{sec:prelim}, we apply identical scoring rules to every source and target graph.

\textbf{Attribute Context.} To link nodes with related content regardless of raw connectivity, we select the $k$ nodes most similar in attribute space to node $i$:
\begin{equation}
\mathcal C_{\Att}(i)=\operatorname{Top}_k\bigl\{\cos(\mathbf x_i,\mathbf x_j):j\in\mathcal V\setminus\{i\}\bigr\}.
\label{eq:att}
\end{equation}
The operator $\operatorname{Top}_k$ returns the highest-scoring indices, keeping all candidates if fewer than $k$ exist. All other nodes are candidates, so isolated nodes and nodes outside the raw neighborhood are compared using the same attribute criterion.

\textbf{Structural Context.} Common raw neighbors define a structural association between nodes with different attributes. Measuring this overlap by $\nu_{ij}=|\mathcal N(i)\cap\mathcal N(j)|$, we select nonadjacent nodes that share at least one common neighbor:
\begin{equation}
\mathcal C_{\Str}(i)=\operatorname{Top}_k\bigl\{\nu_{ij}:j\notin\mathcal N(i)\cup\{i\},\ \nu_{ij}>0\bigr\}.
\label{eq:str}
\end{equation}
These two-hop pairs capture shared neighborhood membership without propagating directly along raw edges.

\textbf{Semantic Context.} We compare neighborhood content even when two nodes have no common neighbors. To capture this association, we compute one-hop summaries $\overline{\mathbf X}=\widehat{\mathbf A}\mathbf X$ and select the $k$ nodes with the most similar neighborhood summaries:
\begin{equation}
\mathcal C_{\Sem}(i)=\operatorname{Top}_k\bigl\{\cos(\overline{\mathbf x}_i,\overline{\mathbf x}_j):j\in\mathcal V\setminus\{i\}\bigr\}.
\label{eq:sem}
\end{equation}
Here, $\widehat{\mathbf A}$ includes self-loops and symmetric normalization. Structural contexts compare neighborhood membership; semantic contexts compare aggregated neighborhood content.

For each type $r\in\mathcal R=\{\Att,\Str,\Sem\}$, we symmetrize the selections into $\mathcal P_r=\{\{i,j\}:j\in\mathcal C_r(i)\text{ or }i\in\mathcal C_r(j)\}$ with binary indicator matrix $\mathbf C_r$, defining the reconstructed propagation pathways. Theorem~\ref{thm:stability} bounds changes in propagation and reconstruction.

\subsection{Multi-Domain Pre-training over Relative Contexts}
\label{sec:pretrain}
\begin{algorithm}[t]
\caption{\recog{} pre-training and target adaptation}
\label{alg:recog}
\fontsize{7.6}{8.9}\selectfont\raggedright
\textbf{Input:} source graphs $\{G_s\}_{s=1}^{M}$; target graph $G_t$ with available supervision; $k_{\mathrm{train}}$, $k$, $\tau$, and update budget $T$.\\
\textbf{Output:} frozen encoder $(f_\phi,g_\psi)$ and a fitted task predictor.
\begin{enumerate}[label=\arabic*:,leftmargin=1.6em,itemsep=0.6pt,topsep=2pt,parsep=0pt]
\item \textbf{for} $s=1,\dots,M$ \textbf{do}
\item \quad Preprocess $G_s$; build $\mathbf C_{s,r}$ ($r\in\mathcal R$) by Eqs.~\eqref{eq:att}--\eqref{eq:sem} with $k_{\mathrm{train}}$
\item \quad $\mathbf S_s\gets\operatorname{Norm}(\mathbf W_s)$ by Eqs.~\eqref{eq:pretrain-graph} and~\eqref{eq:calibration}
\item \textbf{for} $u=1,\dots,T$ \textbf{do}
\item \quad $s\gets1+((u-1)\bmod M)$;\ $\mathbf H_s\gets g_\psi(\mathbf S_s,f_\phi(\mathbf X_s))$
\item \quad Update $\Theta$ by one Adam step on $\mathcal L_s$ in Eq.~\eqref{eq:loss}
\item Freeze $f_\phi$ and $g_\psi$; discard the decoder $D$
\item Preprocess $G_t$; build its contexts with size $k$ by Eqs.~\eqref{eq:att}--\eqref{eq:sem}
\item Select context weights with target supervision (Section~\ref{sec:deploy})
\item Encode $G_t$ by Eq.~\eqref{eq:deploy}; fit the task predictor on $\mathbf H$
\end{enumerate}
\end{algorithm}

We train the shared encoder to aggregate context information while preserving each node's distinctive attributes. SVD preprocessing and a shared multilayer perceptron (MLP) provide fixed-dimensional inputs to graph propagation~\cite{yuan2025much}. Let $P_d$ denote the projection, zero-padding, and row normalization in Section~\ref{sec:prelim}; the MLP $f_\phi$ maps each row of $\mathbf X_s=P_d(\mathbf X_s^{\mathrm{raw}})$ into a common hidden space. During pre-training, we take the union of the three context types, counting each selected pair once with a common within-graph weight $w$:
\begin{equation}
\mathbf W=w\,\mathbf 1\bigl[\textstyle\sum_{r\in\mathcal R}\mathbf C_r>\mathbf 0\bigr],\qquad
\mathbf S=\operatorname{Norm}(\mathbf W).
\label{eq:pretrain-graph}
\end{equation}
The weight $w$ preserves the total weight of a reference graph while distributing it over the context union. With $\mathcal P=\bigcup_r\mathcal P_r$ and $e_+=|\mathcal P\setminus\mathcal E|$, we set
\begin{equation}
\begin{split}
w&=\frac{0.9|\mathcal E|+0.1\beta e_+}{|\mathcal P|},\\
\beta&=\operatorname{clip}_{[0.05,20]}\!\left(\frac{\tau}{1-\tau}\frac{1.8|\mathcal E|+n}{0.2e_+}\right).
\end{split}
\label{eq:calibration}
\end{equation}
Here $\tau=0.8$ is the share of total weight, including self-loops, that newly added pairs receive before $\beta$ is clipped. The operator $\operatorname{Norm}$ adds unit self-loops and applies symmetric normalization. A shared GNN $g_\psi$ then aggregates the projected features over $\mathbf S$:
\begin{equation}
\mathbf H=g_\psi\bigl(\mathbf S,f_\phi(\mathbf X)\bigr).
\label{eq:encoder}
\end{equation}
Both $f_\phi$ and $g_\psi$ share parameters across all source graphs.

Since downstream tasks are unknown during pre-training, we train the encoder via node attribute reconstruction~\cite{hou2022graphmae} to preserve node content after context aggregation. Using $\mathbf X$ as both input and target, we reconstruct node attributes with a shared affine decoder $D(\mathbf h)=\mathbf U\mathbf h+\mathbf b$ by minimizing the cosine loss
\begin{equation}
\mathcal L=\frac1n\sum_{i=1}^{n}(1-\mathbf x_i^\top\widehat{\mathbf x}_i),\quad
\widehat{\mathbf x}_i=\frac{D(\mathbf h_i)}{\|D(\mathbf h_i)\|_\epsilon},
\label{eq:loss}
\end{equation}
where $\|\mathbf z\|_\epsilon=\max(\|\mathbf z\|,\epsilon)$ with $\epsilon>0$ stabilizes normalization. Theorem~\ref{thm:readout} links target reconstruction loss to the classification error of an induced affine readout through the attribute classification margins. We minimize $M^{-1}\sum_{s=1}^{M}\mathcal L_s(\Theta)$ over the shared parameters $\Theta=\{\phi,\psi,\mathbf U,\mathbf b\}$, cycling through sources with one full-graph update per visit (details in Appendix~\ref{app:impl}).

\subsection{Downstream Adaptation with a Frozen Encoder}
\label{sec:deploy}
The useful context types and propagation strengths vary with the target graph and task. We adapt the propagation operator while keeping $f_\phi$ and $g_\psi$ frozen. Applying the same preprocessing and RCR rules to the target graph, we weight each context type $r$ by $s_r=\lambda b_r$, where $b_r\in\{0,1\}$ and the strength $\lambda\ge0$ is shared across types with nonzero weight. Defining $\mathbf Q=\sum_{r\in\mathcal R}b_r\mathbf C_r$, we compute target representations via
\begin{equation}
\mathbf S=\operatorname{Norm}(\Gamma_\lambda(\mathbf Q)),\quad
\mathbf H=g_\psi(\mathbf S,f_\phi(\mathbf X)).
\label{eq:deploy}
\end{equation}
For $q_i=\sum_jQ_{ij}$, let $\alpha_i=\lambda/q_i$ when $q_i>0$ and zero otherwise. The deployment weights are
\begin{equation}
[\Gamma_\lambda(\mathbf Q)]_{ij}=Q_{ij}\sqrt{\alpha_i\alpha_j}.
\label{eq:target-weights}
\end{equation}
This scales the degree-normalized context union before adding self-loops and re-normalizing. At $\lambda=0$, $\mathbf S=\mathbf I$ and each node is encoded using only its own attributes.

\textbf{Adjusted homophily.} For node pairs $\mathcal P$ and node labels, the adjusted homophily~\cite{platonov2023characterizing} is $h_{\mathrm{adj}}(\mathcal P)=\bigl(h-\sum_c \pi_c^2\bigr)/\bigl(1-\sum_c \pi_c^2\bigr)$, where $h$ is the fraction of pairs joining same-class nodes and $\pi_c$ is the empirical frequency of label $c$ across pair endpoints. It equals zero under label independence, and takes positive (negative) values when pairs preferentially link same-class (different-class) nodes.

We select $(b_{\Att},b_{\Str},b_{\Sem},\lambda)$ using available target supervision. With a validation split, we select from a finite candidate grid. In few-shot node classification, because sparse support labels cover few context pairs, a support-set classifier fitted at $\lambda=0$ first generates pseudo-labels to estimate the adjusted homophily of each context type; types with positive adjusted homophily receive nonzero weight, and their estimates set a prior propagation strength (refined via support-set cross-validation at 3 and 5 shots). For standard Q1 classification, a node-specific variant activates all context types and replaces $\lambda$ by $m_i=0.9|\mathcal N(i)|+0.1\beta u_i$, where $u_i$ counts newly added context pairs incident to node $i$; Q2 also includes this variant as a candidate. Other configurations use a shared strength. We discard $D$ and fit task-specific predictors on frozen $\mathbf H$: a linear node classifier, a node-pair predictor, or a pooled graph classifier. Appendix~\ref{app:impl} lists task settings and selection grids.

\setcounter{dbltopnumber}{2}
\renewcommand{\dbltopfraction}{0.95}
\renewcommand{\textfraction}{0.05}
\section{Theoretical Analysis}
\label{sec:theory}
We quantify how differences in attribute mixing create reconstruction conflicts, then bound changes in context-based propagation and the readout error associated with target reconstruction. Supplementary proofs are provided in Appendix~\ref{app:proofs}.

\begin{table*}[t]
\caption{Q1 1-shot node classification accuracy (\%, mean $\pm$ SD) over 500 shared tasks per held-out graph. Avg. denotes the six-target mean. Bold and underlined values mark the best and runner-up.}
\label{tab:fewshot}
\centering
\fontsize{8.3}{9.9}\selectfont
\setlength{\tabcolsep}{2.6pt}
\renewcommand{\arraystretch}{1.0}
\begin{tabular*}{\textwidth}{@{\extracolsep{\fill}}l*{7}{c}@{}}
\toprule
Method & Cora & CiteSeer & PubMed & CS & Photo & Computers & Avg. \\
\midrule
FUG & 46.40\,$\pm$\,8.37 & 36.07\,$\pm$\,6.81 & 52.23\,$\pm$\,9.47 & \underline{78.02}\,$\pm$\,5.88 & 63.62\,$\pm$\,9.77 & \underline{51.27}\,$\pm$\,12.82 & 54.60 \\
TIG & 48.88\,$\pm$\,8.80 & 44.71\,$\pm$\,9.02 & 50.94\,$\pm$\,10.00 & 75.28\,$\pm$\,6.22 & 63.52\,$\pm$\,10.47 & 49.47\,$\pm$\,13.08 & \underline{55.47} \\
GFT & 36.19\,$\pm$\,5.39 & 32.51\,$\pm$\,5.36 & 50.56\,$\pm$\,7.19 & 63.49\,$\pm$\,6.42 & 49.91\,$\pm$\,7.11 & 38.67\,$\pm$\,7.91 & 45.22 \\
MDGPT & 42.49\,$\pm$\,7.94 & 45.13\,$\pm$\,8.81 & \underline{55.34}\,$\pm$\,8.32 & 72.98\,$\pm$\,6.61 & 64.19\,$\pm$\,9.92 & 49.24\,$\pm$\,10.88 & 54.90 \\
SAMGPT & 44.81\,$\pm$\,8.42 & 41.52\,$\pm$\,8.15 & 46.20\,$\pm$\,9.67 & 76.48\,$\pm$\,6.17 & 59.17\,$\pm$\,8.63 & 46.73\,$\pm$\,9.64 & 52.49 \\
MDGFM & 42.43\,$\pm$\,6.05 & 38.84\,$\pm$\,7.54 & 49.73\,$\pm$\,8.60 & 74.11\,$\pm$\,6.55 & \underline{64.24}\,$\pm$\,9.52 & 49.63\,$\pm$\,9.83 & 53.16 \\
BRIDGE & 42.77\,$\pm$\,6.37 & 41.36\,$\pm$\,9.26 & 54.36\,$\pm$\,8.94 & 75.14\,$\pm$\,6.36 & 63.38\,$\pm$\,9.96 & 49.33\,$\pm$\,10.46 & 54.39 \\
MDGMIX & 40.92\,$\pm$\,5.49 & 29.86\,$\pm$\,5.11 & 43.87\,$\pm$\,6.30 & 69.69\,$\pm$\,6.07 & 63.10\,$\pm$\,8.74 & 49.87\,$\pm$\,10.12 & 49.55 \\
\midrule
GraphAny & 39.18\,$\pm$\,7.71 & 35.48\,$\pm$\,7.11 & 51.79\,$\pm$\,8.41 & 64.50\,$\pm$\,8.65 & 55.28\,$\pm$\,11.45 & 43.51\,$\pm$\,12.73 & 48.29 \\
GraphLoRA & \underline{54.41}\,$\pm$\,10.75 & \underline{50.63}\,$\pm$\,14.41 & \textbf{55.59}\,$\pm$\,11.19 & 74.82\,$\pm$\,9.06 & 52.52\,$\pm$\,9.01 & 41.50\,$\pm$\,13.29 & 54.91 \\
\midrule
\textbf{\recog{}} & \textbf{56.47}\,$\pm$\,10.04 & \textbf{53.62}\,$\pm$\,10.14 & 54.15\,$\pm$\,9.58 & \textbf{84.40}\,$\pm$\,5.84 & \textbf{73.25}\,$\pm$\,9.03 & \textbf{56.59}\,$\pm$\,10.37 & \textbf{63.08} \\
\bottomrule
\end{tabular*}
\end{table*}

\begin{figure*}[t]

\centering

\includegraphics[width=\textwidth]{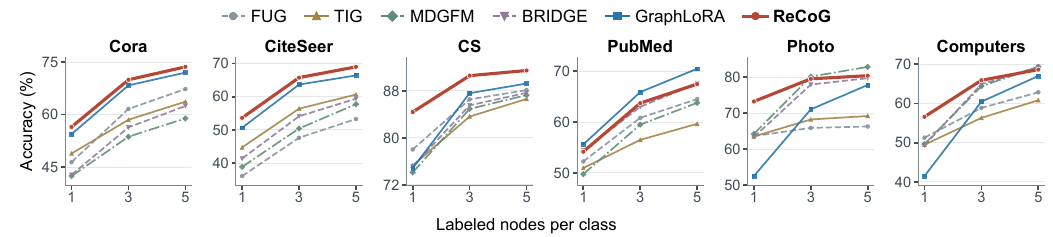}

\caption{Q1 few-shot node classification accuracy (\%) across 1-, 3-, and 5-shot budgets on six held-out target graphs (500 shared tasks per budget). \recog{} is compared against five representative baselines; full results appear in Tables~\ref{tab:fewshot} and~\ref{tab:fewshot-all}.}

\Description{Six panels showing 1-, 3-, and 5-shot node classification accuracy on Cora, CiteSeer, CS, PubMed, Photo, and Computers for FUG, TIG, MDGFM, BRIDGE, GraphLoRA, and ReCoG.}

\label{fig:fewshot-trends}

\end{figure*}

\begin{table*}[t]
\caption{Q2 cross-type transfer accuracy (\%, mean $\pm$ SD) over ten splits. Avg. denotes the six-target mean. Bold and underlined values mark the best and runner-up.}
\label{tab:hetero}
\centering
\fontsize{7.6}{9.0}\selectfont
\setlength{\tabcolsep}{1.2pt}
\renewcommand{\arraystretch}{1.0}
\begin{tabular*}{\textwidth}{@{\extracolsep{\fill}}l*{7}{c}@{}}
\toprule
Method & Cornell & Texas & Wisconsin & Actor & Chameleon & Squirrel & Avg. \\
\midrule
FUG & 48.92\,$\pm$\,8.84 & 62.70\,$\pm$\,4.80 & 56.47\,$\pm$\,3.49 & 28.30\,$\pm$\,1.37 & 35.92\,$\pm$\,2.81 & 26.77\,$\pm$\,1.28 & 43.18 \\
TIG & 47.03\,$\pm$\,7.57 & 58.92\,$\pm$\,3.59 & 50.20\,$\pm$\,5.13 & 27.52\,$\pm$\,1.14 & 36.05\,$\pm$\,1.70 & 26.12\,$\pm$\,0.82 & 40.97 \\
GFT & \underline{71.62}\,$\pm$\,6.65 & \underline{78.11}\,$\pm$\,4.26 & \textbf{83.53}\,$\pm$\,4.49 & \underline{33.57}\,$\pm$\,0.79 & \underline{45.07}\,$\pm$\,1.89 & \underline{35.06}\,$\pm$\,1.02 & \underline{57.83} \\
MDGPT & 61.08\,$\pm$\,8.13 & 62.97\,$\pm$\,2.97 & 55.29\,$\pm$\,3.37 & 28.44\,$\pm$\,0.86 & 40.86\,$\pm$\,1.87 & 33.13\,$\pm$\,1.35 & 46.96 \\
SAMGPT & 60.54\,$\pm$\,6.07 & 62.43\,$\pm$\,4.43 & 56.86\,$\pm$\,4.21 & 28.86\,$\pm$\,0.88 & 42.11\,$\pm$\,3.34 & 32.98\,$\pm$\,0.99 & 47.30 \\
MDGFM & 61.62\,$\pm$\,7.72 & 62.97\,$\pm$\,5.14 & 55.88\,$\pm$\,7.14 & 28.96\,$\pm$\,1.25 & 42.65\,$\pm$\,2.75 & 32.90\,$\pm$\,0.97 & 47.50 \\
BRIDGE & 61.62\,$\pm$\,8.44 & 62.43\,$\pm$\,4.09 & 55.29\,$\pm$\,4.71 & 28.62\,$\pm$\,0.72 & 42.70\,$\pm$\,2.22 & 31.67\,$\pm$\,1.35 & 47.05 \\
MDGMIX & 51.89\,$\pm$\,5.51 & 57.84\,$\pm$\,7.07 & 50.20\,$\pm$\,3.84 & 26.32\,$\pm$\,1.17 & 34.71\,$\pm$\,3.28 & 28.38\,$\pm$\,1.19 & 41.56 \\
\midrule
\textbf{\recog{}} & \textbf{74.59}\,$\pm$\,5.01 & \textbf{80.00}\,$\pm$\,7.37 & \underline{82.94}\,$\pm$\,5.04 & \textbf{34.31}\,$\pm$\,0.90 & \textbf{60.02}\,$\pm$\,3.17 & \textbf{41.91}\,$\pm$\,1.46 & \textbf{62.30} \\
\bottomrule
\end{tabular*}
\end{table*}

\begin{table*}[t]
\caption{Q3 transfer from ogbn-products: 5-shot node classification, link prediction, and graph classification. Entries are accuracy or AUC (\%, mean$_{\pm\mathrm{SD}}$); each Avg. weights three targets equally. Bold and underlined values mark the best and runner-up.}
\label{tab:q3-products}
\centering
\fontsize{7.2}{8.6}\selectfont
\setlength{\tabcolsep}{1.2pt}
\renewcommand{\arraystretch}{1.0}
\begin{tabular*}{\textwidth}{@{\extracolsep{\fill}}l*{12}{c}@{}}
\toprule
& \multicolumn{4}{c}{Node classification (Acc.)} & \multicolumn{4}{c}{Link prediction (AUC)} & \multicolumn{4}{c}{Graph classification (Acc.)} \\
\cmidrule(lr){2-5}\cmidrule(lr){6-9}\cmidrule(lr){10-13}
Method & Cora & CiteS. & PubM. & Avg. & Phys. & Wiki. & DBLP & Avg. & IMDB & COLL. & PROT. & Avg. \\
\midrule
FUG & 66.15$_{\pm4.59}$ & 58.84$_{\pm3.59}$ & 64.97$_{\pm5.44}$ & 63.32 & 98.02$_{\pm0.09}$ & 98.74$_{\pm0.07}$ & 92.50$_{\pm0.13}$ & 96.42 & 68.26$_{\pm2.16}$ & 75.76$_{\pm0.96}$ & 71.14$_{\pm3.24}$ & 71.72 \\
TIG & \underline{66.53}$_{\pm3.94}$ & \underline{61.04}$_{\pm3.25}$ & 63.24$_{\pm4.68}$ & \underline{63.60} & \textbf{98.54}$_{\pm0.05}$ & 91.32$_{\pm0.35}$ & \underline{93.73}$_{\pm0.18}$ & 94.53 & 69.64$_{\pm1.84}$ & \textbf{76.30}$_{\pm1.36}$ & 72.01$_{\pm2.23}$ & 72.65 \\
GFT & 38.76$_{\pm3.54}$ & 37.23$_{\pm2.85}$ & 54.09$_{\pm4.71}$ & 43.36 & 98.02$_{\pm0.04}$ & 98.54$_{\pm0.03}$ & 87.13$_{\pm0.39}$ & 94.56 & 70.05$_{\pm2.84}$ & 75.59$_{\pm1.26}$ & \underline{72.42}$_{\pm1.63}$ & \underline{72.69} \\
MDGPT & 60.03$_{\pm3.79}$ & 59.74$_{\pm3.37}$ & 64.09$_{\pm4.53}$ & 61.29 & 98.24$_{\pm0.03}$ & \underline{99.01}$_{\pm0.04}$ & 92.25$_{\pm0.39}$ & 96.50 & 69.32$_{\pm1.58}$ & 73.30$_{\pm1.27}$ & 71.82$_{\pm1.96}$ & 71.48 \\
SAMGPT & 59.80$_{\pm4.36}$ & 59.44$_{\pm3.53}$ & 56.41$_{\pm7.00}$ & 58.55 & 98.25$_{\pm0.04}$ & 98.95$_{\pm0.06}$ & 91.14$_{\pm0.30}$ & 96.11 & \textbf{70.77}$_{\pm1.52}$ & 74.29$_{\pm1.06}$ & 72.41$_{\pm1.16}$ & 72.49 \\
MDGFM & 63.27$_{\pm3.59}$ & 48.02$_{\pm3.79}$ & 51.05$_{\pm6.44}$ & 54.11 & 98.23$_{\pm0.02}$ & \textbf{99.02}$_{\pm0.05}$ & 91.54$_{\pm0.35}$ & 96.26 & 69.32$_{\pm1.79}$ & 73.43$_{\pm1.12}$ & 71.77$_{\pm1.91}$ & 71.51 \\
BRIDGE & 63.12$_{\pm3.45}$ & 58.61$_{\pm3.35}$ & \textbf{67.75}$_{\pm4.61}$ & 63.16 & 98.32$_{\pm0.04}$ & \underline{99.01}$_{\pm0.04}$ & 92.26$_{\pm0.36}$ & \underline{96.53} & 69.29$_{\pm1.82}$ & 73.69$_{\pm1.07}$ & 71.74$_{\pm1.46}$ & 71.57 \\
\midrule
\textbf{\recog{}} & \textbf{74.63}$_{\pm2.58}$ & \textbf{67.92}$_{\pm2.82}$ & \underline{67.66}$_{\pm4.98}$ & \textbf{70.07} & \underline{98.51}$_{\pm0.05}$ & 98.85$_{\pm0.05}$ & \textbf{94.95}$_{\pm0.08}$ & \textbf{97.44} & \underline{70.13}$_{\pm1.87}$ & \underline{76.04}$_{\pm1.09}$ & \textbf{73.82}$_{\pm1.42}$ & \textbf{73.33} \\
\bottomrule
\end{tabular*}
\end{table*}

\textbf{Shared reconstruction under different graph mixing.} Following two-class analyses of graph convolution~\cite{baranwal2021graphconv,wu2023oversmoothing}, consider two equal-sized node groups carrying unit attributes $\mathbf x_+$ and $\mathbf x_-$ at angle $\vartheta\in(0,\pi)$, mapped by a shared nodewise encoder to distinct prototypes $\mathbf f_+,\mathbf f_-$. On graph $s$, suppose each propagation step assigns mass $q_s$ to the opposite group. Two linear aggregation steps yield
\[
\mathbf z_{s,\pm}=\bm\mu_f\pm t_s\bm\nu_f,\quad
\bm\mu_f=\tfrac12(\mathbf f_++\mathbf f_-),\quad
\bm\nu_f=\tfrac12(\mathbf f_+-\mathbf f_-),
\]
where $t_s=(1-2q_s)^2>0$ is the surviving attribute contrast. Let $\ell_{s,\pm}$ be the clipped cosine loss in Eq.~\eqref{eq:loss}.

\begin{proposition}[Cost of shared reconstruction]
\label{prop:shared-cost}
Write $t_{\min}=\min_s t_s$, $t_{\max}=\max_s t_s$, $\chi=(t_{\max}-t_{\min})/(t_{\max}+t_{\min})$, and $\delta_* =\frac12\arcsin(\chi\sin\vartheta)$. Each graph admits a zero-loss affine decoder, whereas one shared affine decoder satisfies
\begin{equation}
\inf_D\max_{s,\pm}\ell_{s,\pm}(D)=1-\cos\delta_*.
\label{eq:cosine-shared-cost}
\end{equation}
For $N$ equally weighted graphs, the mean loss is at least $(1-\cos\delta_*)/(2N)$.
\end{proposition}
\begin{proof}
An affine decoder independently chooses its value at $\bm\mu_f$ and its action on $\bm\nu_f\ne0$, so each graph alone has zero loss. For $t_{\min}<t_{\max}$, put $r=t_{\min}/t_{\max}$, $\alpha=(1+r)/2$, and $\beta=(1-r)/2$. The outputs on the two endpoint graphs obey
\[
\mathbf Y_{\min,\pm}=\alpha\mathbf Y_{\max,\pm}+\beta\mathbf Y_{\max,\mp},
\qquad \beta/\alpha=\chi.
\]
Set $\omega=\vartheta/2$. If every loss were below $1-\cos\delta_*$, all four output directions would have target angles bounded by some $\delta<\delta_*<\min\{\omega,\pi/2-\omega\}$. Projection onto the target plane preserves this bound. In coordinates where $\mathbf x_\pm=(\cos\omega,\pm\sin\omega)$, write the projected $t_{\max}$-outputs as $a(1,x)$ and $b(1,-y)$, with $a,b>0$. Their slopes satisfy $A\le x,y\le B$, where $A=\tan(\omega-\delta)$ and $B=\tan(\omega+\delta)$. The two mixed outputs must lie in the same target cones, implying
\[
\alpha a(x-A)\ge\beta b(y+A),\qquad
\alpha b(y-A)\ge\beta a(x+A).
\]
Multiplication gives
\[
\chi^2\le\frac{(x-A)(y-A)}{(x+A)(y+A)}
\le\left(\frac{B-A}{B+A}\right)^2
=\frac{\sin^2(2\delta)}{\sin^2\vartheta},
\]
contradicting $\sin(2\delta_*)=\chi\sin\vartheta$.
For attainment, choose the affine decoder with outputs $\lambda(1,\pm\kappa t_s)$ in this plane, where $\kappa=\tan(\omega+\delta_*)/t_{\max}$ and $\lambda\ge\epsilon$. The identity $\tan(\omega-\delta_*)/\tan(\omega+\delta_*)=t_{\min}/t_{\max}$ bounds every target angle by $\delta_*$, with equality at the endpoints. All norms exceed $\epsilon$, so clipping is inactive. For the lower bound, clipping only increases positive-alignment losses. The case $t_{\min}=t_{\max}$ has zero shared loss, and the mean bound follows because at least one of the $2N$ group losses reaches the minimax bound.
\end{proof}

In this two-group construction, varying mixing rates ($t_{\min}<t_{\max}$) impose conflicting reconstruction demands even when node attributes are identical across graphs. If a nonlinear encoder achieves exact reconstruction under a shared decoder, it must correct this mismatch in its representations: for any nonzero decoded outputs $\mathbf Y_{s,\pm}$ with linear mixture $\mathbf B_\pm=\alpha\mathbf Y_{\max,\pm}+\beta\mathbf Y_{\max,\mp}$, exact reconstruction requires relative deviation $\rho=\max_\pm\|\mathbf Y_{\min,\pm}-\mathbf B_\pm\|/\|\mathbf Y_{\min,\pm}\|$ satisfying (Appendix~\ref{app:proofs})
\begin{equation}
\rho\ge\frac{\chi\sin\vartheta}{\sqrt{1+2\chi\cos\vartheta+\chi^2}}>0.
\label{eq:zero-error-compensation}
\end{equation}
Since $\mathbf Y_{\min,\pm}-\mathbf B_\pm=\mathbf U(\mathbf h_{\min,\pm}-\alpha\mathbf h_{\max,\pm}-\beta\mathbf h_{\max,\mp})$ under an affine decoder, the learned representations must supply this correction.

Relative contexts remove this mismatch in a family of regular graphs (Appendix~\ref{app:proofs}): given two graphs with $m_0$ nodes per group—one wired within groups and the other across groups—with $k=m_0-1$, attribute and semantic contexts select all within-group pairs while structural contexts exclude cross-group pairs. Both context unions form the same two within-group cliques, preserving the attribute prototypes and admitting a shared zero-loss decoder.

\textbf{Stability of context-based propagation.} We next examine how changes in reconstructed contexts affect a fixed model. For symmetric nonnegative context weights $\mathbf W_s,\mathbf W_t$ on the same node set, define the maximum relative incident-weight change, normalized by degree including self-loops, as
\[
\eta=\max_i\frac{\sum_j|[\mathbf W_t-\mathbf W_s]_{ij}|}{1+\sum_j[\mathbf W_s]_{ij}}.
\]

\begin{theorem}[Propagation and reconstruction stability]
\label{thm:stability}
The normalized operators satisfy
\begin{equation}
\|\operatorname{Norm}(\mathbf W_t)-\operatorname{Norm}(\mathbf W_s)\|_2
\le2\min\{1,\eta\}.
\label{eq:operator-stability}
\end{equation}
For attribute row norms at most one, a fixed model that combines a nodewise Lipschitz map, a two-layer GCN of Eq.~\eqref{eq:gcn} with ReLU, and an affine decoder satisfies
\begin{equation}
|\mathcal L_s(\Theta)-\mathcal L_t(\Theta)|
\le\kappa_{\Theta,\epsilon}\bigl(\delta_X+2\min\{1,\eta\}\bigr),
\label{eq:loss-stability}
\end{equation}
where $\delta_X=\|\mathbf X_s-\mathbf X_t\|_F/\sqrt n$ with fixed-model constant $\kappa_{\Theta,\epsilon}$.
\end{theorem}

Theorem~\ref{thm:stability} bounds operator and reconstruction changes with graph-size-independent constants. RCR links raw-edge changes to this bound through candidate eligibility and top-$k$ rankings. With attributes fixed and consistent tie-breaking, attribute contexts remain unchanged. Structural or semantic selections change only when candidate eligibility changes or a candidate crosses the top-$k$ cutoff. For unchanged candidate sets, a cutoff gap greater than twice the maximum score perturbation preserves the selected sets: no excluded score can overtake a selected score. When these selections and the calibration weight $w$ are preserved, $\eta=0$ and the propagation operator is identical. Propagation therefore depends on changes in the selected relative associations and their weights, rather than on each raw-edge change.

\textbf{Reconstruction supports a linear readout.} Let $\mu(\mathbf x,y)=\mathbf v_y^\top\mathbf x-\max_{y'\ne y}\mathbf v_{y'}^\top\mathbf x$ be the margin of linear attribute class scores, let $\eta_{\rm cls}(\gamma)$ denote the fraction of target nodes with margin at most $\gamma$, and put $B_v=\max_{y\ne y'}\|\mathbf v_y-\mathbf v_{y'}\|$.

\begin{theorem}[Reconstruction bounds the readout error]
\label{thm:readout}
For attribute row norms at most one and $D(\mathbf h)=\mathbf U\mathbf h+\mathbf b$, the readout $g(\mathbf h)=\arg\max_y\mathbf v_y^\top D(\mathbf h)$ has target misclassification fraction
\[
R(g)\le\eta_{\rm cls}(\gamma)+2B_v^2\mathcal L_t/\gamma^2,\qquad\gamma>0.
\]
\end{theorem}

Accurate target reconstruction preserves attribute-based class distinctions in an affine readout of the frozen representations: when most nodes have margins above a fixed $\gamma>0$, small $\mathcal L_t$ yields a low misclassification bound.

\section{Experiments}
\label{sec:exp}
We evaluate \recog{} across 18 target datasets to answer three questions: \textbf{Q1:} Does it transfer to unseen graph domains across different label budgets? \textbf{Q2:} Does it transfer from homophilous source graphs to heterophilous target graphs? \textbf{Q3:} Does pre-training on a single large graph support node-, edge-, and graph-level tasks with a frozen encoder? We then analyze context weighting, context construction, and the pre-training objective, followed by test-time robustness and hyperparameter sensitivity.

\subsection{Experimental Setup}
\label{sec:setup}
\textbf{Datasets.} Q1 uses six homophilous benchmarks: the citation networks Cora, CiteSeer, and PubMed~\cite{sen2008collective,yang2016revisiting}, the co-authorship network CS, and the co-purchase networks Photo and Computers~\cite{shchur2018pitfalls}. Q2 uses these six graphs as sources and transfers to six heterophilous targets: Cornell, Texas, Wisconsin, Actor~\cite{pei2020geom}, Chameleon, and Squirrel~\cite{rozemberczki2021multi}. Q3 pre-trains on either ogbn-products or ogbn-arxiv~\cite{hu2020ogb} and evaluates on nine targets: Cora, CiteSeer, and PubMed for node classification; Physics~\cite{shchur2018pitfalls}, WikiCS~\cite{mernyei2020wiki}, and DBLP~\cite{bojchevski2018deep} for link prediction; and IMDB-B, COLLAB, and PROTEINS~\cite{morris2020tudataset} for graph classification. Dataset statistics and task settings are summarized in Appendix~\ref{app:impl}.

\textbf{Baselines.} We compare \recog{} with eight multi-domain graph pre-training methods: FUG~\cite{zhao2024fug}, TIG~\cite{zhao2026towards}, GFT~\cite{wang2024gft}, MDGPT~\cite{yu2024text}, SAMGPT~\cite{yu2025samgpt}, MDGFM~\cite{wang2025multidomain}, BRIDGE~\cite{yuan2025much}, and MDGMIX~\cite{mdgmix2026}. For few-shot transfer, GraphAny~\cite{zhao2025fully} uses training-free target inference, while GraphLoRA~\cite{yang2025graphlora} adapts through low-rank updates. Results of three graph self-supervised methods, DGI~\cite{velickovic2019dgi}, GRACE~\cite{zhu2020deep}, and BGRL~\cite{thakoor2022largescale}, are reported in Appendix~\ref{app:additional-results}. We use official implementations and input preprocessing. We evaluate each baseline with its official adaptation procedure or with the shared frozen-representation evaluation in Appendix~\ref{app:impl}.

\textbf{Evaluation.} \recog{} uses one shared set of pre-training hyperparameters across all source collections and keeps its encoder frozen on target graphs (Appendix~\ref{app:impl}). For few-shot node classification, all methods use the same 500 sampled tasks per target at 1, 3, and 5 shots. Standard node classification uses three random 20\%/80\% labeled/test splits, and we select the $\ell_2$ regularization strength of the logistic regression classifier by cross-validation within the labeled set. Q2 uses the ten public splits with linear evaluation, and Q3 uses frozen encoders with shared readouts on nodes, node pairs (evaluated by AUC), or pooled graph representations (evaluated by accuracy). Target test labels are excluded from model and context-weight selection.

\subsection{Cross-Domain Transfer (Q1)}
\label{sec:q1}
We pre-train on five unlabeled Q1 graphs and evaluate on the held-out sixth graph. We report 1-shot accuracy in Table~\ref{tab:fewshot}, per-target trajectories across 1, 3, and 5 shots in Figure~\ref{fig:fewshot-trends}, and the 3-/5-shot results in Table~\ref{tab:fewshot-all} (Appendix~\ref{app:additional-results}). The margin of the six-target mean over the strongest baseline narrows from $7.61$ percentage points (pp) at 1 shot to $3.07$ and $1.27$~pp at 3 and 5 shots, respectively. This trend indicates that the frozen representations are most useful when labeled target nodes are scarce. In the 1-shot setting, \recog{} ranks first on five targets across citation, co-authorship, and co-purchase domains, outperforming the runner-up by $6.38$~pp on CS and $9.01$~pp on Photo.

\recog{} remains first on Cora, CiteSeer, and CS at both 3 and 5 shots, while GraphLoRA leads on PubMed and MDGFM leads on Photo. Under the standard 20\%-labeled protocol, \recog{} reaches $87.44\%$ mean accuracy and ranks first on five targets (Table~\ref{tab:nc}, Appendix~\ref{app:additional-results}), indicating that its advantage extends beyond the few-shot setting.

\begin{figure}[t]

\centering

\includegraphics[width=1.0\linewidth,trim=0 2bp 0 1bp,clip]{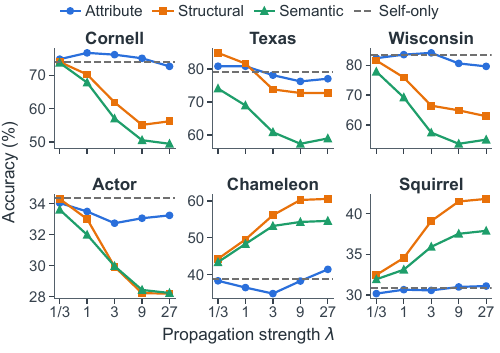}

\caption{Q2 accuracy (\%) by context type and strength $\lambda$. Dashed: self-only encoding ($\lambda=0$).}

\Description{Six panels showing how individual context types respond to propagation strength on Cornell, Texas, Wisconsin, Actor, Chameleon, and Squirrel.}

\label{fig:strength}

\end{figure}

\begin{figure*}[t]

\centering

\includegraphics[width=\textwidth]{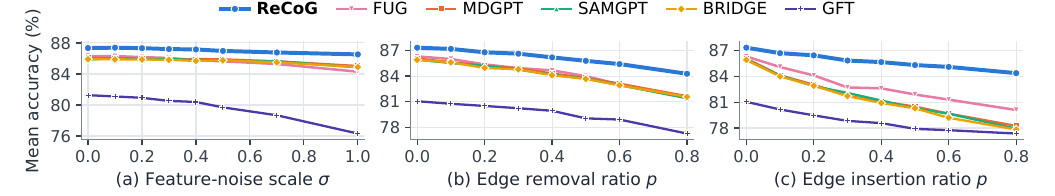}

\caption{Test-time robustness: (a) Gaussian feature noise, (b) edge removal, and (c) edge insertion. Mean accuracy (\%) over the six Q1 graphs, with rebuilt contexts and refitted classifiers on frozen representations.}

\Description{Three line-plot panels comparing six-graph mean accuracy of six methods under feature noise, edge removal, and edge insertion.}

\label{fig:robust}

\end{figure*}

\subsection{Cross-Type Transfer (Q2)}
\label{sec:q2}
To assess cross-type transfer, we pre-train on all six homophilous Q1 graphs and evaluate on six heterophilous targets. In Table~\ref{tab:hetero}, \recog{} ranks first on five targets and reaches $62.30\%$ mean accuracy---surpassing the strongest baseline (GFT, $57.83\%$) by $4.47$~pp, with gains over GFT of $14.95$~pp on Chameleon and $6.85$~pp on Squirrel, while GFT leads on Wisconsin.

Figure~\ref{fig:strength} shows how accuracy changes with the propagation strength of each context type. Cornell and Wisconsin favor attribute contexts, Texas favors weak structural propagation, and Chameleon and Squirrel benefit from stronger structural propagation. Actor favors minimal or self-only propagation. The contrast between Texas and the two Wikipedia graphs indicates that the preferred strength of structural propagation varies across targets. Selecting context types and strengths accommodates these differences while the encoder remains frozen.

\subsection{Cross-Task Transfer (Q3)}
\label{sec:q3}
We pre-train one encoder on ogbn-products and evaluate its frozen representations on nine targets spanning three tasks. In Table~\ref{tab:q3-products}, \recog{} achieves the highest task-wise mean in every task: $70.07\%$ accuracy for 5-shot node classification ($+6.47$~pp over the strongest baseline), $97.44\%$ AUC for link prediction ($+0.91$~pp), and $73.33\%$ accuracy for graph classification ($+0.64$~pp). The largest gains occur in node classification, where \recog{} exceeds the runner-up by $8.10$~pp on Cora and $6.88$~pp on CiteSeer.

With ogbn-arxiv as the source, ReCoG again leads on Cora, CiteSeer, DBLP, and PROTEINS, with task-wise means of $68.55\%$, $97.20\%$, and $73.32\%$ (Table~\ref{tab:q3}, Appendix~\ref{app:additional-results}). Results from both sources show that one frozen encoder supports node-, edge-, and graph-level tasks.

\subsection{Model Analysis}
\label{sec:analysis}
\begin{table}[t]
\caption{Q1 1-shot ablation accuracy (\%, mean$_{\pm\mathrm{SD}}$) over 500 shared tasks per target. $\Delta$Avg.\ is the six-target mean change (pp) from full \recog{} (bold; 63.08\% mean).}
\label{tab:ablation}
\centering
\fontsize{6.6}{8.2}\selectfont
\setlength{\tabcolsep}{0.35pt}
\renewcommand{\arraystretch}{1.50}
\begin{tabular*}{\columnwidth}{@{\extracolsep{\fill}}l*{6}{c}r@{}}
\toprule
Variant & Cora & CiteS. & PubM. & CS & Photo & Comp. & $\Delta$Avg. \\
\midrule
\textbf{Full \recog{}} & \textbf{56.47}$_{\pm10.04}$ & \textbf{53.62}$_{\pm10.14}$ & \textbf{54.15}$_{\pm9.58}$ & \textbf{84.40}$_{\pm5.84}$ & \textbf{73.25}$_{\pm9.03}$ & \textbf{56.59}$_{\pm10.37}$ & -- \\
\midrule
w/o Attribute & 58.35$_{\pm10.14}$ & 53.10$_{\pm10.17}$ & 54.46$_{\pm9.90}$ & 83.48$_{\pm6.23}$ & 70.68$_{\pm9.22}$ & 54.73$_{\pm10.29}$ & $-0.61$ \\
w/o Structural & 52.66$_{\pm9.80}$ & 52.33$_{\pm9.98}$ & 52.03$_{\pm9.07}$ & 83.17$_{\pm6.18}$ & 71.40$_{\pm9.05}$ & 55.14$_{\pm10.17}$ & $-1.96$ \\
w/o Semantic & 49.74$_{\pm9.62}$ & 47.63$_{\pm9.42}$ & 53.95$_{\pm9.89}$ & 83.23$_{\pm6.06}$ & 72.15$_{\pm8.67}$ & 55.89$_{\pm10.56}$ & $-2.65$ \\
Raw edges & 34.95$_{\pm7.39}$ & 36.44$_{\pm7.43}$ & 49.84$_{\pm8.58}$ & 79.72$_{\pm5.75}$ & 65.77$_{\pm9.96}$ & 49.90$_{\pm10.46}$ & $-10.31$ \\
\midrule
Contrastive & 48.57$_{\pm8.27}$ & 30.59$_{\pm5.14}$ & 41.21$_{\pm6.37}$ & 77.48$_{\pm6.59}$ & 68.11$_{\pm8.33}$ & 56.79$_{\pm10.27}$ & $-9.29$ \\
Link recon. & 54.49$_{\pm8.93}$ & 47.63$_{\pm8.91}$ & 53.04$_{\pm8.53}$ & 83.30$_{\pm6.67}$ & 70.35$_{\pm8.27}$ & 57.66$_{\pm10.07}$ & $-2.00$ \\
\bottomrule
\end{tabular*}
\end{table}

\textbf{Context construction.} Table~\ref{tab:ablation} reports 1-shot ablation accuracy on six Q1 targets over the same 500 tasks. Replacing relative contexts with raw edges during pre-training and adaptation lowers accuracy on all six targets, reducing mean accuracy by $10.31$~pp ($-21.52$~pp on Cora and $-17.18$~pp on CiteSeer). Removing structural contexts lowers accuracy on all six targets ($-1.96$~pp on average), indicating that common-neighbor relations consistently improve transfer. Removing semantic contexts causes the largest drops on Cora ($-6.73$~pp) and CiteSeer ($-5.99$~pp), suggesting that these citation graphs benefit most from neighborhood attribute summaries. Attribute contexts improve accuracy on Photo ($+2.57$~pp) and Computers ($+1.86$~pp) but slightly reduce it on Cora and PubMed, which supports selecting context types for each target.

\textbf{Pre-training objective.} Attribute reconstruction achieves the highest accuracy among the three pre-training objectives on five targets. Replacing it with contrastive learning or link reconstruction reduces the six-target mean by $9.29$~pp and $2.00$~pp, respectively. Contrastive learning causes a drop of $23.03$~pp on CiteSeer, whereas link reconstruction slightly outperforms attribute reconstruction on Computers. These results indicate that training aggregated representations to retain node attributes benefits frozen transfer, consistent with the readout bound in Theorem~\ref{thm:readout}.

\textbf{Source connectivity.}\label{app:source-edge} Removing 90\% of source edges before RCR pre-training, while keeping target graphs and sampled tasks fixed, changes mean 1-shot accuracy by only $-0.17$~pp, from $63.08\%$ to $62.91\%$. This repeats the source-side intervention of Figure~\ref{fig:source_edge_deletion} on \recog{}. In contrast, replacing relative contexts with raw-edge propagation lowers accuracy by $10.31$~pp. Relative contexts therefore remain effective after substantial source-edge removal.

\textbf{Test-time robustness.}\label{sec:robustness}
We pre-train on the six clean Q1 graphs and perturb those same graphs at test time, rebuilding contexts and refitting classifiers with the encoder frozen. Figure~\ref{fig:robust} reports mean accuracy under Gaussian feature noise, edge removal, and edge insertion. ReCoG leads throughout the tested ranges; under the strongest perturbation, accuracy decreases by $0.82$, $3.06$, and $2.96$~pp, respectively. Attribute contexts retain their selection rule under edge perturbations, while structural and semantic contexts respond to neighborhood changes. This complements the source-edge intervention above, which evaluates transfer to clean unseen targets. Appendix~\ref{app:robustness} specifies the perturbations.

\textbf{Hyperparameter sensitivity.}\label{app:sensitivity}
Figure~\ref{fig:sensitivity} varies target $k$ from 6 to 32 at $\tau=0.8$, and $\tau$ from 0.5 to 0.9 at $k=12$, with $k_{\mathrm{train}}=16$. Averaged over three random splits of the standard Q1 protocol, mean accuracy ranges from $87.05\%$ to $87.44\%$ as $k$ varies and from $87.35\%$ to $87.44\%$ as $\tau$ varies, and thus remains stable. Here, $\tau$ controls both pre-training calibration and node-specific target strengths.

\section{Conclusion}
\label{sec:conclusion}
We study how the raw graph structure affects multi-domain graph pre-training. Removing a large proportion of edges from source graphs improves cross-domain transfer for four representative methods, and our analysis traces this effect to graph-dependent attribute mixing, which imposes conflicting reconstruction demands on a shared model. We therefore propose ReCoG, which constructs attribute, structural, and semantic contexts under comparison rules shared across graphs, pre-trains a single encoder by attribute reconstruction over them, and adapts the frozen encoder by reweighting context types. Across 18 target datasets, ReCoG leads in mean cross-domain and cross-type accuracy and supports node-, edge-, and graph-level prediction.

\bibliographystyle{ACM-Reference-Format}
\bibliography{references}

\setcounter{dbltopnumber}{2}
\renewcommand{\dbltopfraction}{0.7}
\renewcommand{\dblfloatpagefraction}{0.95}
\renewcommand{\textfraction}{0.05}
\appendix
\raggedbottom
\section{Implementation and Evaluation}
\label{app:impl}\label{app:datasets}\label{app:baselines}\label{app:coverage}\label{app:protocols}

\textbf{Datasets.} Table~\ref{tab:datasets} summarizes the 18 targets and two additional sources. Q3 node classification reuses Cora, CiteSeer, and PubMed. IMDB-B and COLLAB use one-hot node degrees. Other inputs follow the cited dataset releases~\cite{sen2008collective,yang2016revisiting,shchur2018pitfalls,pei2020geom,rozemberczki2021multi,hu2020ogb,mernyei2020wiki,bojchevski2018deep,morris2020tudataset}.
\begin{table}[b]
\caption{Dataset statistics. Labels of link-prediction targets are unused (--); IMDB-B and COLLAB use one-hot node degrees (deg.).}
\label{tab:datasets}
\centering\fontsize{7.0}{8.1}\selectfont
\setlength{\tabcolsep}{2.4pt}
\begin{tabular}{lrrrr}\toprule
Dataset & Nodes & Edges & Feat. & Cls.\\\midrule
Cora & 2,708 & 5,278 & 1,433 & 7\\
CiteSeer & 3,327 & 4,552 & 3,703 & 6\\
PubMed & 19,717 & 44,324 & 500 & 3\\
CS & 18,333 & 81,894 & 6,805 & 15\\
Photo & 7,650 & 119,081 & 745 & 8\\
Computers & 13,752 & 245,861 & 767 & 10\\\midrule
Cornell & 183 & 277 & 1,703 & 5\\
Texas & 183 & 279 & 1,703 & 5\\
Wisconsin & 251 & 450 & 1,703 & 5\\
Actor & 7,600 & 26,659 & 932 & 5\\
Chameleon & 2,277 & 31,371 & 2,325 & 5\\
Squirrel & 5,201 & 198,353 & 2,089 & 5\\\midrule
ogbn-products & 2,449,029 & 61,859,140 & 100 & 47\\
ogbn-arxiv & 169,343 & 1,166,243 & 128 & 40\\
Physics & 34,493 & 247,962 & 8,415 & --\\
WikiCS & 11,701 & 216,123 & 300 & --\\
DBLP & 17,716 & 52,867 & 1,639 & --\\\midrule
Dataset & Graphs & Mean nodes & Feat. & Cls.\\
IMDB-B & 1,000 & 19.8 & deg. & 2\\
COLLAB & 5,000 & 74.5 & deg. & 3\\
PROTEINS & 1,113 & 39.1 & 3 & 2\\\bottomrule
\end{tabular}
\end{table}

\begin{table*}[t]
\caption{Q1 accuracy (\%, mean$_{\pm\mathrm{SD}}$) at 3 and 5 shots over 500 shared tasks per target. Avg.\ weights six targets equally; 1-shot results appear in Table~\ref{tab:fewshot}. Bold and underlined values mark the best and runner-up.}
\label{tab:fewshot-all}
\centering
\fontsize{7.2}{8.6}\selectfont
\setlength{\tabcolsep}{0.9pt}
\renewcommand{\arraystretch}{1.0}
\begin{tabular*}{\textwidth}{@{\extracolsep{\fill}}l*{14}{c}@{}}
\toprule
& \multicolumn{2}{c}{Cora} & \multicolumn{2}{c}{CiteSeer} & \multicolumn{2}{c}{PubMed} & \multicolumn{2}{c}{CS} & \multicolumn{2}{c}{Photo} & \multicolumn{2}{c}{Computers} & \multicolumn{2}{c}{Avg.} \\
\cmidrule(lr){2-3}\cmidrule(lr){4-5}\cmidrule(lr){6-7}\cmidrule(lr){8-9}\cmidrule(lr){10-11}\cmidrule(lr){12-13}\cmidrule(lr){14-15}
Method & 3 & 5 & 3 & 5 & 3 & 5 & 3 & 5 & 3 & 5 & 3 & 5 & 3 & 5 \\
\midrule
DGI & 45.56$_{\pm5.08}$ & 53.07$_{\pm3.79}$ & 44.96$_{\pm4.42}$ & 50.55$_{\pm3.54}$ & 55.58$_{\pm5.12}$ & 60.34$_{\pm4.47}$ & 85.26$_{\pm2.08}$ & 87.77$_{\pm1.20}$ & 70.17$_{\pm5.23}$ & 74.10$_{\pm4.33}$ & 56.98$_{\pm7.27}$ & 63.16$_{\pm4.49}$ & 59.75 & 64.83 \\
GRACE & 38.59$_{\pm4.00}$ & 44.50$_{\pm3.25}$ & 30.63$_{\pm3.88}$ & 34.72$_{\pm3.42}$ & 45.68$_{\pm4.85}$ & 49.95$_{\pm4.23}$ & 65.49$_{\pm3.51}$ & 71.96$_{\pm2.75}$ & 57.22$_{\pm6.52}$ & 64.24$_{\pm5.42}$ & 50.64$_{\pm6.08}$ & 56.71$_{\pm5.15}$ & 48.04 & 53.68 \\
BGRL & 46.74$_{\pm4.09}$ & 53.28$_{\pm3.37}$ & 31.66$_{\pm3.58}$ & 35.74$_{\pm3.11}$ & 51.06$_{\pm5.51}$ & 56.07$_{\pm4.68}$ & 75.12$_{\pm3.29}$ & 81.12$_{\pm2.22}$ & 72.63$_{\pm7.23}$ & 77.83$_{\pm5.18}$ & 62.05$_{\pm8.43}$ & 67.13$_{\pm6.08}$ & 56.54 & 61.86 \\
\midrule
FUG & 61.63$_{\pm5.42}$ & 67.31$_{\pm3.23}$ & 47.62$_{\pm4.94}$ & 53.28$_{\pm3.86}$ & 60.82$_{\pm6.32}$ & 64.51$_{\pm4.64}$ & 86.55$_{\pm1.70}$ & 88.13$_{\pm1.13}$ & 65.91$_{\pm8.90}$ & 66.29$_{\pm7.50}$ & 58.85$_{\pm10.17}$ & 62.83$_{\pm8.18}$ & 63.56 & 67.06 \\
TIG & 58.53$_{\pm6.17}$ & 63.70$_{\pm4.59}$ & 56.39$_{\pm5.03}$ & 60.66$_{\pm3.45}$ & 56.48$_{\pm7.81}$ & 59.64$_{\pm6.58}$ & 83.63$_{\pm3.04}$ & 86.62$_{\pm1.98}$ & 68.24$_{\pm10.30}$ & 69.20$_{\pm8.48}$ & 56.31$_{\pm11.03}$ & 60.80$_{\pm8.49}$ & 63.26 & 66.77 \\
GFT & 52.88$_{\pm4.54}$ & 60.01$_{\pm3.61}$ & 44.98$_{\pm4.53}$ & 50.91$_{\pm3.80}$ & 60.66$_{\pm5.86}$ & 65.06$_{\pm4.36}$ & 78.95$_{\pm2.74}$ & 82.69$_{\pm1.62}$ & 68.41$_{\pm5.63}$ & 73.61$_{\pm5.07}$ & 54.17$_{\pm6.17}$ & 59.97$_{\pm5.41}$ & 60.01 & 65.38 \\
MDGPT & 54.99$_{\pm6.30}$ & 60.86$_{\pm3.97}$ & 57.34$_{\pm4.97}$ & 61.87$_{\pm3.28}$ & \underline{63.88}$_{\pm6.02}$ & 67.59$_{\pm4.71}$ & 82.70$_{\pm2.45}$ & 85.35$_{\pm1.51}$ & 76.31$_{\pm5.62}$ & 78.20$_{\pm4.92}$ & 63.10$_{\pm8.20}$ & 67.17$_{\pm6.31}$ & 66.39 & 70.17 \\
SAMGPT & 58.66$_{\pm6.25}$ & 64.52$_{\pm3.90}$ & 54.44$_{\pm6.27}$ & 60.74$_{\pm3.98}$ & 52.99$_{\pm9.13}$ & 57.45$_{\pm8.61}$ & 85.27$_{\pm2.00}$ & 87.01$_{\pm1.24}$ & 74.87$_{\pm5.26}$ & 78.14$_{\pm3.81}$ & 61.08$_{\pm6.77}$ & 66.69$_{\pm5.11}$ & 64.55 & 69.09 \\
MDGFM & 53.68$_{\pm4.89}$ & 58.91$_{\pm3.70}$ & 50.41$_{\pm4.89}$ & 57.75$_{\pm5.34}$ & 59.48$_{\pm6.25}$ & 63.78$_{\pm4.98}$ & 84.93$_{\pm2.17}$ & 87.28$_{\pm1.30}$ & \textbf{80.15}$_{\pm4.43}$ & \textbf{82.86}$_{\pm3.67}$ & 64.38$_{\pm7.07}$ & \textbf{69.30}$_{\pm5.39}$ & 65.50 & 69.98 \\
BRIDGE & 56.36$_{\pm5.15}$ & 62.45$_{\pm3.43}$ & 54.10$_{\pm6.57}$ & 59.38$_{\pm4.86}$ & 62.98$_{\pm6.00}$ & \underline{67.64}$_{\pm4.69}$ & 85.53$_{\pm1.96}$ & 87.68$_{\pm1.28}$ & 77.94$_{\pm5.44}$ & 79.72$_{\pm4.71}$ & \underline{64.87}$_{\pm7.39}$ & \underline{69.29}$_{\pm5.41}$ & 66.96 & 71.03 \\
MDGMIX & 53.55$_{\pm4.54}$ & 59.39$_{\pm3.44}$ & 39.56$_{\pm4.29}$ & 44.87$_{\pm3.77}$ & 52.03$_{\pm5.22}$ & 57.39$_{\pm4.55}$ & 83.40$_{\pm2.49}$ & 86.66$_{\pm1.46}$ & 76.90$_{\pm4.83}$ & 80.72$_{\pm3.68}$ & 63.03$_{\pm6.84}$ & 68.09$_{\pm4.66}$ & 61.41 & 66.19 \\
\midrule
GraphAny & 57.17$_{\pm5.93}$ & 64.86$_{\pm4.42}$ & 50.15$_{\pm5.53}$ & 56.42$_{\pm4.13}$ & 61.89$_{\pm6.18}$ & 66.39$_{\pm4.95}$ & 82.90$_{\pm3.73}$ & 86.73$_{\pm2.00}$ & 77.05$_{\pm6.39}$ & \underline{82.30}$_{\pm4.45}$ & 62.53$_{\pm8.05}$ & 67.97$_{\pm6.20}$ & 65.28 & 70.78 \\
GraphLoRA & \underline{68.37}$_{\pm5.29}$ & \underline{72.02}$_{\pm3.35}$ & \underline{63.67}$_{\pm7.58}$ & \underline{66.42}$_{\pm4.08}$ & \textbf{65.89}$_{\pm7.76}$ & \textbf{70.48}$_{\pm5.45}$ & \underline{87.61}$_{\pm2.47}$ & \underline{89.25}$_{\pm1.48}$ & 71.06$_{\pm6.59}$ & 77.82$_{\pm4.89}$ & 60.44$_{\pm8.67}$ & 66.97$_{\pm6.68}$ & \underline{69.51} & \underline{73.83} \\
\midrule
\textbf{\recog{}} & \textbf{69.96}$_{\pm5.25}$ & \textbf{73.65}$_{\pm2.68}$ & \textbf{65.76}$_{\pm4.71}$ & \textbf{69.00}$_{\pm2.65}$ & 63.71$_{\pm6.34}$ & 67.47$_{\pm4.62}$ & \textbf{90.59}$_{\pm1.49}$ & \textbf{91.47}$_{\pm0.98}$ & \underline{79.54}$_{\pm5.53}$ & 80.41$_{\pm4.75}$ & \textbf{65.89}$_{\pm6.38}$ & 68.60$_{\pm4.11}$ & \textbf{72.58} & \textbf{75.10} \\
\bottomrule
\end{tabular*}
\end{table*}

\begin{table*}[t]
\caption{Standard Q1 classification accuracy (\%, mean$_{\pm\mathrm{SD}}$) over three 20\%/80\% labeled/test splits. Avg. denotes the six-target mean. Bold and underlined values mark the best and runner-up.}
\label{tab:nc}
\centering
\fontsize{7.2}{8.6}\selectfont
\setlength{\tabcolsep}{1.2pt}
\renewcommand{\arraystretch}{1.0}
\begin{tabular*}{\textwidth}{@{\extracolsep{\fill}}l*{7}{c}@{}}
\toprule
Method & Cora & CiteSeer & PubMed & CS & Photo & Computers & Avg. \\
\midrule
DGI & 75.47$_{\pm0.78}$ & 65.76$_{\pm1.19}$ & 82.83$_{\pm0.24}$ & 92.82$_{\pm0.18}$ & 91.48$_{\pm0.26}$ & 87.24$_{\pm0.21}$ & 82.60 \\
GRACE & 72.00$_{\pm0.36}$ & 56.29$_{\pm0.71}$ & 76.80$_{\pm0.37}$ & 90.29$_{\pm0.09}$ & 90.76$_{\pm0.15}$ & 86.82$_{\pm0.15}$ & 78.83 \\
BGRL & 79.43$_{\pm1.24}$ & 64.46$_{\pm1.14}$ & 83.91$_{\pm0.26}$ & 92.63$_{\pm0.21}$ & 92.76$_{\pm0.21}$ & 88.98$_{\pm0.16}$ & 83.70 \\
\midrule
FUG & 82.79$_{\pm0.72}$ & 70.52$_{\pm0.15}$ & 84.42$_{\pm0.23}$ & 92.41$_{\pm0.02}$ & 92.31$_{\pm0.16}$ & 87.30$_{\pm0.16}$ & 84.96 \\
TIG & \underline{84.39}$_{\pm1.08}$ & \underline{73.08}$_{\pm0.52}$ & \textbf{86.45}$_{\pm0.13}$ & \underline{93.42}$_{\pm0.05}$ & 93.27$_{\pm0.22}$ & 89.51$_{\pm0.07}$ & \underline{86.69} \\
GFT & 72.99$_{\pm0.94}$ & 64.79$_{\pm0.99}$ & 83.59$_{\pm0.12}$ & 93.06$_{\pm0.27}$ & 90.25$_{\pm0.18}$ & 84.18$_{\pm0.19}$ & 81.48 \\
MDGPT & 81.78$_{\pm0.47}$ & 72.25$_{\pm0.49}$ & \underline{85.76}$_{\pm0.24}$ & 93.02$_{\pm0.08}$ & 93.27$_{\pm0.07}$ & 90.01$_{\pm0.09}$ & 86.02 \\
SAMGPT & 81.64$_{\pm0.56}$ & 71.59$_{\pm0.34}$ & 85.44$_{\pm0.23}$ & 93.04$_{\pm0.09}$ & 92.98$_{\pm0.20}$ & 89.91$_{\pm0.16}$ & 85.77 \\
MDGFM & 81.60$_{\pm0.70}$ & 71.66$_{\pm0.61}$ & 85.63$_{\pm0.25}$ & 92.84$_{\pm0.09}$ & 93.33$_{\pm0.15}$ & \underline{90.19}$_{\pm0.13}$ & 85.88 \\
BRIDGE & 81.59$_{\pm0.65}$ & 71.36$_{\pm0.51}$ & 85.47$_{\pm0.23}$ & 93.11$_{\pm0.05}$ & 93.20$_{\pm0.13}$ & 90.07$_{\pm0.20}$ & 85.80 \\
MDGMIX & 80.74$_{\pm1.21}$ & 67.24$_{\pm1.04}$ & 83.80$_{\pm0.27}$ & 93.02$_{\pm0.01}$ & \underline{93.37}$_{\pm0.22}$ & 89.89$_{\pm0.07}$ & 84.68 \\
\midrule
\textbf{\recog{}} & \textbf{84.86}$_{\pm0.17}$ & \textbf{74.07}$_{\pm0.09}$ & 84.84$_{\pm0.19}$ & \textbf{94.67}$_{\pm0.06}$ & \textbf{94.99}$_{\pm0.25}$ & \textbf{91.23}$_{\pm0.09}$ & \textbf{87.44} \\
\bottomrule
\end{tabular*}
\end{table*}

\begin{table*}[t]
\caption{Q3 transfer from ogbn-arxiv under the same protocol as Table~\ref{tab:q3-products}. Entries are accuracy or AUC (\%, mean$_{\pm\mathrm{SD}}$); each Avg. weights three targets equally. Bold and underlined values mark the best and runner-up.}
\label{tab:q3}
\centering
\fontsize{7.2}{8.6}\selectfont
\setlength{\tabcolsep}{1.2pt}
\renewcommand{\arraystretch}{1.0}
\begin{tabular*}{\textwidth}{@{\extracolsep{\fill}}l*{12}{c}@{}}
\toprule
& \multicolumn{4}{c}{Node classification (Acc.)} & \multicolumn{4}{c}{Link prediction (AUC)} & \multicolumn{4}{c}{Graph classification (Acc.)} \\
\cmidrule(lr){2-5}\cmidrule(lr){6-9}\cmidrule(lr){10-13}
Method & Cora & CiteS. & PubM. & Avg. & Phys. & Wiki. & DBLP & Avg. & IMDB & COLL. & PROT. & Avg. \\
\midrule
DGI & 52.14$_{\pm3.25}$ & 46.08$_{\pm3.40}$ & 56.76$_{\pm4.35}$ & 51.66 & 98.00$_{\pm0.03}$ & 98.81$_{\pm0.04}$ & 91.15$_{\pm0.29}$ & 95.99 & 69.62$_{\pm2.01}$ & 75.82$_{\pm1.04}$ & 71.20$_{\pm2.41}$ & 72.21 \\
GRACE & 47.83$_{\pm3.33}$ & 33.69$_{\pm3.08}$ & 51.26$_{\pm4.37}$ & 44.26 & 97.67$_{\pm0.03}$ & 98.72$_{\pm0.04}$ & 90.36$_{\pm0.41}$ & 95.58 & 68.87$_{\pm2.26}$ & 74.72$_{\pm1.17}$ & \underline{73.06}$_{\pm1.16}$ & 72.22 \\
BGRL & 56.29$_{\pm3.56}$ & 36.06$_{\pm3.09}$ & 55.54$_{\pm4.58}$ & 49.30 & \underline{98.46}$_{\pm0.03}$ & 98.95$_{\pm0.05}$ & 91.77$_{\pm0.18}$ & 96.39 & 69.11$_{\pm2.35}$ & 75.33$_{\pm1.47}$ & 72.49$_{\pm1.76}$ & 72.31 \\
\midrule
FUG & 57.19$_{\pm5.58}$ & 46.79$_{\pm4.51}$ & 64.03$_{\pm5.30}$ & 56.00 & 98.15$_{\pm0.06}$ & 98.71$_{\pm0.06}$ & 92.46$_{\pm0.19}$ & 96.44 & 68.45$_{\pm2.41}$ & 75.53$_{\pm1.06}$ & 71.78$_{\pm2.65}$ & 71.92 \\
TIG & \underline{65.87}$_{\pm4.01}$ & 60.74$_{\pm3.28}$ & 62.03$_{\pm4.75}$ & 62.88 & \textbf{98.56}$_{\pm0.05}$ & 90.62$_{\pm0.28}$ & \underline{93.82}$_{\pm0.15}$ & 94.33 & 68.94$_{\pm2.38}$ & \textbf{76.34}$_{\pm1.60}$ & 72.47$_{\pm2.52}$ & \underline{72.58} \\
GFT & 35.78$_{\pm3.73}$ & 30.15$_{\pm2.71}$ & 56.26$_{\pm4.23}$ & 40.73 & 97.65$_{\pm0.04}$ & 98.50$_{\pm0.03}$ & 85.25$_{\pm0.39}$ & 93.80 & 69.12$_{\pm2.16}$ & 75.21$_{\pm1.47}$ & 72.54$_{\pm1.98}$ & 72.29 \\
MDGPT & 60.48$_{\pm3.70}$ & \underline{63.41}$_{\pm2.96}$ & 65.92$_{\pm5.00}$ & \underline{63.27} & 98.20$_{\pm0.05}$ & 98.97$_{\pm0.05}$ & 92.04$_{\pm0.33}$ & 96.40 & 68.22$_{\pm2.04}$ & 73.47$_{\pm1.17}$ & 71.79$_{\pm1.72}$ & 71.16 \\
SAMGPT & 60.40$_{\pm5.93}$ & 59.49$_{\pm3.70}$ & 59.03$_{\pm7.09}$ & 59.64 & 98.16$_{\pm0.03}$ & 98.93$_{\pm0.04}$ & 91.28$_{\pm0.53}$ & 96.12 & 69.91$_{\pm1.74}$ & 73.08$_{\pm1.19}$ & 72.33$_{\pm1.57}$ & 71.77 \\
MDGFM & 65.52$_{\pm3.16}$ & 56.78$_{\pm3.39}$ & 60.83$_{\pm5.65}$ & 61.04 & 98.30$_{\pm0.03}$ & \underline{99.00}$_{\pm0.04}$ & 91.06$_{\pm0.51}$ & 96.12 & 69.66$_{\pm1.87}$ & 73.12$_{\pm1.40}$ & 72.12$_{\pm2.31}$ & 71.63 \\
BRIDGE & 63.24$_{\pm3.45}$ & 58.70$_{\pm3.33}$ & \textbf{66.89}$_{\pm4.72}$ & 62.94 & 98.31$_{\pm0.04}$ & \textbf{99.01}$_{\pm0.05}$ & 92.29$_{\pm0.36}$ & \underline{96.54} & \underline{70.01}$_{\pm1.38}$ & 73.79$_{\pm0.99}$ & 71.79$_{\pm1.14}$ & 71.86 \\
\midrule
\textbf{\recog{}} & \textbf{71.61}$_{\pm2.74}$ & \textbf{67.58}$_{\pm3.09}$ & \underline{66.45}$_{\pm4.91}$ & \textbf{68.55} & 98.28$_{\pm0.04}$ & 98.76$_{\pm0.03}$ & \textbf{94.56}$_{\pm0.14}$ & \textbf{97.20} & \textbf{70.02}$_{\pm1.78}$ & \underline{76.02}$_{\pm1.02}$ & \textbf{73.91}$_{\pm1.62}$ & \textbf{73.32} \\
\bottomrule
\end{tabular*}
\end{table*}

\begin{figure*}[t]

\centering

\includegraphics[width=0.72\textwidth]{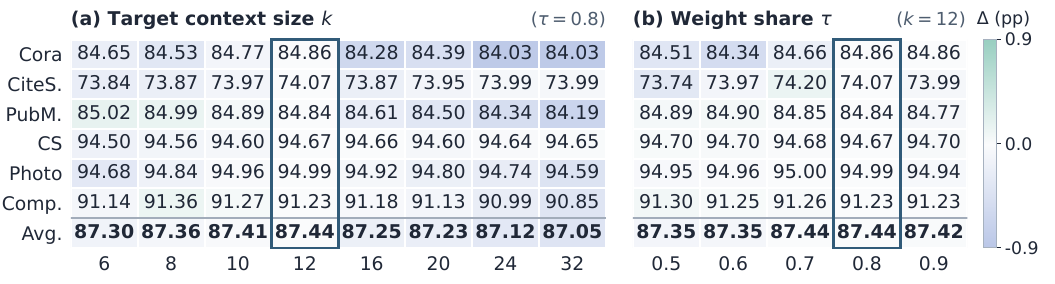}

\caption{Standard Q1 hyperparameter sensitivity ($k_{\mathrm{train}}=16$): (a) target context size $k$ and (b) context weight share $\tau$. Cells report accuracy (\%) over three splits; colors show shifts ($\Delta$, pp) from the outlined defaults ($k=12$, $\tau=0.8$).}

\Description{Two annotated heatmaps showing per-dataset accuracy and average accuracy across target context sizes and context weight shares.}

\label{fig:sensitivity}

\end{figure*}

\textbf{Pre-training.} We set $d=512$, using a nodewise MLP $f_\phi$ with hidden widths $1024$ and $512$ followed by a linear layer, a two-layer GNN $g_\psi$ with widths $256$ and $512$ and ReLU, an affine decoder $D$, and $\epsilon=10^{-8}$. Each context selects $k_{\mathrm{train}}=16$ candidates; a fixed random key breaks common-neighbor ties. Adam uses learning rate $10^{-4}$, weight decay $10^{-3}$, a cosine schedule, gradient clipping at $10$, seed $0$, and $1{,}000$ updates cycling through the sources. These settings are shared across all source collections. Context construction uses blocked similarity searches on $\mathbf X$ and $\widehat{\mathbf A}\mathbf X$, and the sparse product $\mathbf A^2$, on one GPU.

\textbf{Context selection.} Pre-training and deployment weights follow Eqs.~\eqref{eq:calibration} and~\eqref{eq:target-weights}. Standard Q1 uses these node-specific strengths and selects $k\in\{6,8,10,12,16,20,24\}$ on a 70/30 holdout of the labeled set ($k=12$). Q2 uses $k=32$ and compares $\{\Att\},\{\Str\},\{\Sem\},\{\Str,\Sem\},\{\Att,\Str,\Sem\}$ at $\lambda\in\{1/3,1,3,9,27\}$, plus self-only ($\lambda=0$) and node-specific propagation (27 candidates), on each split's validation nodes; each candidate requires one forward pass of the frozen encoder.

\textbf{Few-shot adaptation.} With $k=32$, a logistic classifier fitted to the support nodes at $\lambda=0$ supplies pseudo-labels. We select context types with positive adjusted homophily $\hat h_r$ and round $\sum_{r:\hat h_r>0}\bar d_r\hat h_r$ ($\bar d_r$ denoting the mean degree of type $r$) to the nearest strength on the log grid above. At 3 or 5 shots, leave-one-out support validation compares the grid strengths and replaces the prior choice only if a one-sided McNemar test favors the validation-best strength, i.e., $b-c\ge\max(1,1.645\sqrt{b+c})$, where $b,c$ count support nodes classified correctly only by the validation-best or prior choice, respectively. Every method uses the same $500$ tasks per target and budget; all remaining nodes are test nodes, and Q3 node classification uses 5 shots. Reported task SDs measure variation across the $500$ tasks.

\textbf{Readouts and splits.} Standard Q1 uses three 20\%/80\% labeled/test splits and $\ell_2$-normalized representations. Logistic regression selects $C\in\{2^{-10},\ldots,2^{10}\}$ using five 50/50 shuffle splits inside the labeled set, then refits on all labeled nodes (staying inside the 70\% training portion when selecting $k$). Q2 uses the ten public train/validation/test masks. Link prediction uses five 85\%/5\%/10\% edge splits, using only training edges for encoding and context construction; a shared two-layer 256-unit pair MLP uses $[\mathbf z_u,\mathbf z_v,|\mathbf z_u-\mathbf z_v|,\mathbf z_u\odot\mathbf z_v]$, binary cross-entropy, equal-sized fresh random negatives, and validation-AUC early stopping. Graph classification uses TUDataset attributes or one-hot degrees, standardized mean/max pooled representations, and a two-layer 256-unit MLP with dropout $0.5$ across 50 stratified 20\%/80\% splits with inner validation stopping. Link and graph tasks use all context types with $k=\min(32,\lceil\sqrt n\rceil)$ and $\lambda=9$ and $1$, respectively.

\textbf{Baselines.} Pre-trained baselines use the same source collections and their official preprocessing: DGI, GRACE, and BGRL use SVD-projected attributes, whereas FUG and TIG use their native feature encoders. For downstream adaptation, MDGPT, SAMGPT, MDGFM, and BRIDGE tune target prompts with frozen encoders and select epochs by support loss; GFT fine-tunes with its prototype and linear objectives. GraphLoRA uses low-rank adaptation, and GraphAny performs training-free inference. Frozen-representation evaluations share these task readouts. Averages weight targets equally: six for Q1/Q2 and three per Q3 task.

\section{Additional Results}
\label{app:additional-results}\label{app:analysis}

Tables~\ref{tab:fewshot-all}--\ref{tab:q3} and Figure~\ref{fig:sensitivity} give additional Q1 and Q3 results.

\textbf{Perturbation protocol.}\label{app:robustness}
For Figure~\ref{fig:robust}, we add independent Gaussian noise with scale $\sigma\operatorname{std}(\mathbf X)$, remove a fraction $p$ of undirected edges, or insert $p|\mathcal E|$ random non-edges. All encoders are jointly pre-trained on the six clean Q1 graphs and evaluated on their corrupted versions using the three-repeat 20\%/80\% protocol. We recompute preprocessing under feature noise, rebuild contexts, and refit classifiers while freezing each encoder.

\section{Supplementary Proofs}
\label{app:proofs}
Proposition~\ref{prop:shared-cost} is proved in the main text. Below we prove its nonlinear consequence, the context construction, and both theorems.

\textbf{Nonlinear correction, Eq.~\eqref{eq:zero-error-compensation}.}
Zero clipped cosine loss implies $\mathbf Y_{s,\pm}=c_{s,\pm}\mathbf x_\pm$ with $c_{s,\pm}\ge\epsilon$. Choose the sign whose opposite $t_{\max}$-graph coefficient is no smaller than its own. The angle between its mixture $\mathbf B_\pm$ and target is at least
\[
b_0=\arctan\frac{\chi\sin\vartheta}{1+\chi\cos\vartheta}\in(0,\pi/2).
\]
Indeed, that mixture is proportional to $\mathbf x_\pm+z\mathbf x_\mp$ with $z\ge\chi$, and its angle increases with $z$. For a vector at angle $b$ to a target ray, minimizing $\|a\mathbf x-\mathbf B\|/a$ over $a>0$ gives $\sin b$ for $b<\pi/2$, and an infimum of one otherwise. Therefore $\rho\ge\sin b_0$, giving the stated bound. For the actual affine decoder, the bias cancels because $\alpha+\beta=1$, yielding the representation identity in the main text. \qed

\textbf{Relative-context construction.}
Take $d\ge2$, $m_0\ge d+2$, and $m_0d$ even. On two groups of $m_0$ nodes, use a $d$-regular graph within each group for $G_1$ and a $d$-regular bipartite graph for $G_2$. Their raw contrast factors are $t_1=1$ and $t_2=((d-1)/(d+1))^2$. Set $k=m_0-1$. Attribute contexts select all within-group pairs. The aggregated attributes of $G_2$ are proportional to $\mathbf x_++d\mathbf x_-$ and $d\mathbf x_++\mathbf x_-$; these have distinct directions, so semantic contexts also select within-group pairs. Common neighbors imply same-group membership in either graph. The union is two cliques; every positive uniform calibrated weight preserves the constant group prototypes under normalization. A shared affine decoder recovers both targets exactly. The condition $m_0\ge d+2$ ensures newly added pairs exist for calibration. The construction uses two attribute directions and is preserved by a common orthogonal SVD basis and zero-padding. \qed

\textbf{Propagation bound in Theorem~\ref{thm:stability}.}
Let $\mathbf E=\mathbf W_s-\mathbf W_t$, $d_{a,i}=1+\sum_j[\mathbf W_a]_{ij}$, $\bm\Delta_a=\operatorname{diag}(d_{a,i})$, and $\mathbf R=\bm\Delta_t^{1/2}\bm\Delta_s^{-1/2}$. Then
\[
\mathbf S_s-\mathbf S_t
=\bm\Delta_s^{-1/2}\mathbf E\bm\Delta_s^{-1/2}
+\mathbf R\mathbf S_t\mathbf R-\mathbf S_t.
\]
The first symmetric term is similar to $\bm\Delta_s^{-1}\mathbf E$, whose absolute row sums are at most $\eta$, so its norm is at most $\eta$. Degree changes obey $|d_{t,i}-d_{s,i}|\le\eta d_{s,i}$. Hence the diagonal entries $r_i$ of $\mathbf R$ satisfy $|r_ir_j-1|\le\eta$. Entrywise nonnegativity gives
\[
|\mathbf x^\top(\mathbf R\mathbf S_t\mathbf R-\mathbf S_t)\mathbf x|
\le\eta|\mathbf x|^\top\mathbf S_t|\mathbf x|
\le\eta\|\mathbf x\|^2.
\]
Here $\|\mathbf S_a\|_2=1$ by similarity to a stochastic matrix. The triangle inequality then gives the bounds $2\eta$ and $2$, proving Eq.~\eqref{eq:operator-stability}. \qed

\textbf{Reconstruction bound in Theorem~\ref{thm:stability}.}
Write $\mathbf F_a=f_\phi(\mathbf X_a)$ and $\mathbf Y_a=\operatorname{ReLU}(\mathbf S_a\mathbf F_a\bm\Theta_1+\mathbf1\mathbf b_1^\top)$, with $\mathbf H_a=\mathbf S_a\mathbf Y_a\bm\Theta_2+\mathbf1\mathbf b_2^\top$. Let $L_j=\|\bm\Theta_j\|_2$, $L_f$ be the nodewise Lipschitz constant, and $\delta_S=\|\mathbf S_s-\mathbf S_t\|_2$. Nonexpansiveness gives
\[
\begin{aligned}
\|\mathbf Y_s-\mathbf Y_t\|_F&\le L_1\delta_S\|\mathbf F_s\|_F\\
&\quad+L_1L_f\|\mathbf X_s-\mathbf X_t\|_F,\\
\|\mathbf H_s-\mathbf H_t\|_F&\le L_2(\delta_S\|\mathbf Y_s\|_F+\|\mathbf Y_s-\mathbf Y_t\|_F).
\end{aligned}
\]
Put $C_f=L_f+\|f_\phi(0)\|$. The unit-row constraint gives $\|\mathbf F_s\|_F\le\sqrt n C_f$ and $\|\mathbf Y_s\|_F\le\sqrt n(L_1C_f+\|\mathbf b_1\|)$. Clipped normalization is projection onto the unit ball after division by $\epsilon$, and is thus $1/\epsilon$-Lipschitz. With $c=\|\mathbf U\|_2L_2/\epsilon$, Cauchy--Schwarz yields
\[
|\mathcal L_s-\mathcal L_t|
\le(1+cL_1L_f)\delta_X+c(2L_1C_f+\|\mathbf b_1\|)\delta_S.
\]
Taking $\kappa_{\Theta,\epsilon}$ as the larger coefficient and using Eq.~\eqref{eq:operator-stability} proves the claim. A second ReLU preserves the bound. \qed

\textbf{Readout bound in Theorem~\ref{thm:readout}.}
Let $\widehat{\mathbf x}$ be the clipped normalized decoder output. Both row norms are at most one, giving
\[
\|\widehat{\mathbf x}-\mathbf x\|^2\le2(1-\mathbf x^\top\widehat{\mathbf x}).
\]
Positive normalization preserves class-score ordering. If $\mu(\mathbf x,y)>\gamma$ but $g$ selects $y'\ne y$, then
\[
\gamma<(\mathbf v_y-\mathbf v_{y'})^\top(\mathbf x-\widehat{\mathbf x})
\le B_v\|\mathbf x-\widehat{\mathbf x}\|.
\]
Thus the error indicator is at most $\mathbf1[\mu\le\gamma]+2B_v^2(1-\mathbf x^\top\widehat{\mathbf x})/\gamma^2$. Averaging over target nodes proves the theorem. \qed

\end{document}